\documentclass[journal]{IEEEtran}

\usepackage{amsmath,amssymb,amsfonts}
\usepackage{graphicx}
\usepackage{cite}
\usepackage{booktabs}
\usepackage{microtype}
\usepackage{multirow}
\usepackage[table]{xcolor}
\usepackage{url}
\usepackage[colorlinks=false,
            pdfborder={0 0 0},
            pdfauthor={Jie Xu, Kangjin Yu, Ziyi Jin, Junjie Gao, Liqing Chen, Yixian Li, Shuai Tian, Zhongpu Xia},
            pdftitle={JEPA Policy: Diffusion-Free Imitation Learning via Paired Action and Future Representation Prediction},
            pdfsubject={Diffusion-free robot imitation learning},
            pdfkeywords={imitation learning, joint-embedding predictive architecture, predictive representation learning, robot learning, robotic manipulation}]{hyperref}

\begin{document}

\title{JEPA Policy: Diffusion-Free Imitation Learning via Paired Action and Future Representation Prediction}

\author{Jie~Xu, Kangjin~Yu, Ziyi~Jin, Junjie~Gao, Liqing~Chen, Yixian~Li,
Shuai~Tian, and Zhongpu~Xia\textsuperscript{*}\\[0.5ex]
\normalfont\normalsize Anyverse Dynamics\thanks{Jie Xu: \href{mailto:jeff_xu_0503@foxmail.com}{jeff\_xu\_0503@foxmail.com}. \textsuperscript{*}Corresponding author: Zhongpu Xia.}}

\markboth{Preprint \quad \href{https://github.com/jiejie567/JEPA-Policy}{Code: github.com/jiejie567/JEPA-Policy}}{J. Xu \MakeLowercase{\textit{et al.}}: JEPA Policy}

\bstctlcite{IEEEtran:BSTcontrol}
\maketitle

\begin{abstract}
Standard behavior cloning supervises actions without explicitly constraining the future representation paired with each demonstrated action chunk. We introduce JEPA Policy, a diffusion-free framework that uses the action chunk and its observed future representation as paired training targets. Action and future-representation tokens interact in a shared Transformer and are refined through two forward passes. Future prediction can therefore shape the representation used to generate actions. Dual-branch and gradient-routing controls attribute the gain to this shared topology rather than to an auxiliary prediction head alone. Across nine simulated tasks, JEPA Policy improves mean success over the action-only MIP baseline and outperforms Diffusion Policy under the evaluated configurations, while adding $0.29$\,ms to MIP's model latency. A five-task, $630$-episode physical-robot study produces the same pooled ranking. Further audits find no complete representation collapse under action supervision and identify a task-conditioned failure-ranking signal in future-prediction error. These results support paired future-representation supervision as a practical approach to low-latency visuomotor imitation without iterative generative sampling.
\end{abstract}

{\small\noindent
\textbf{Code:} \url{https://github.com/jiejie567/JEPA-Policy}\par
\noindent\textbf{Project page:} \url{https://jiejie567.github.io/JEPA-Policy/}\par}

\begin{figure}[!t]
\centering
\includegraphics[width=\columnwidth]{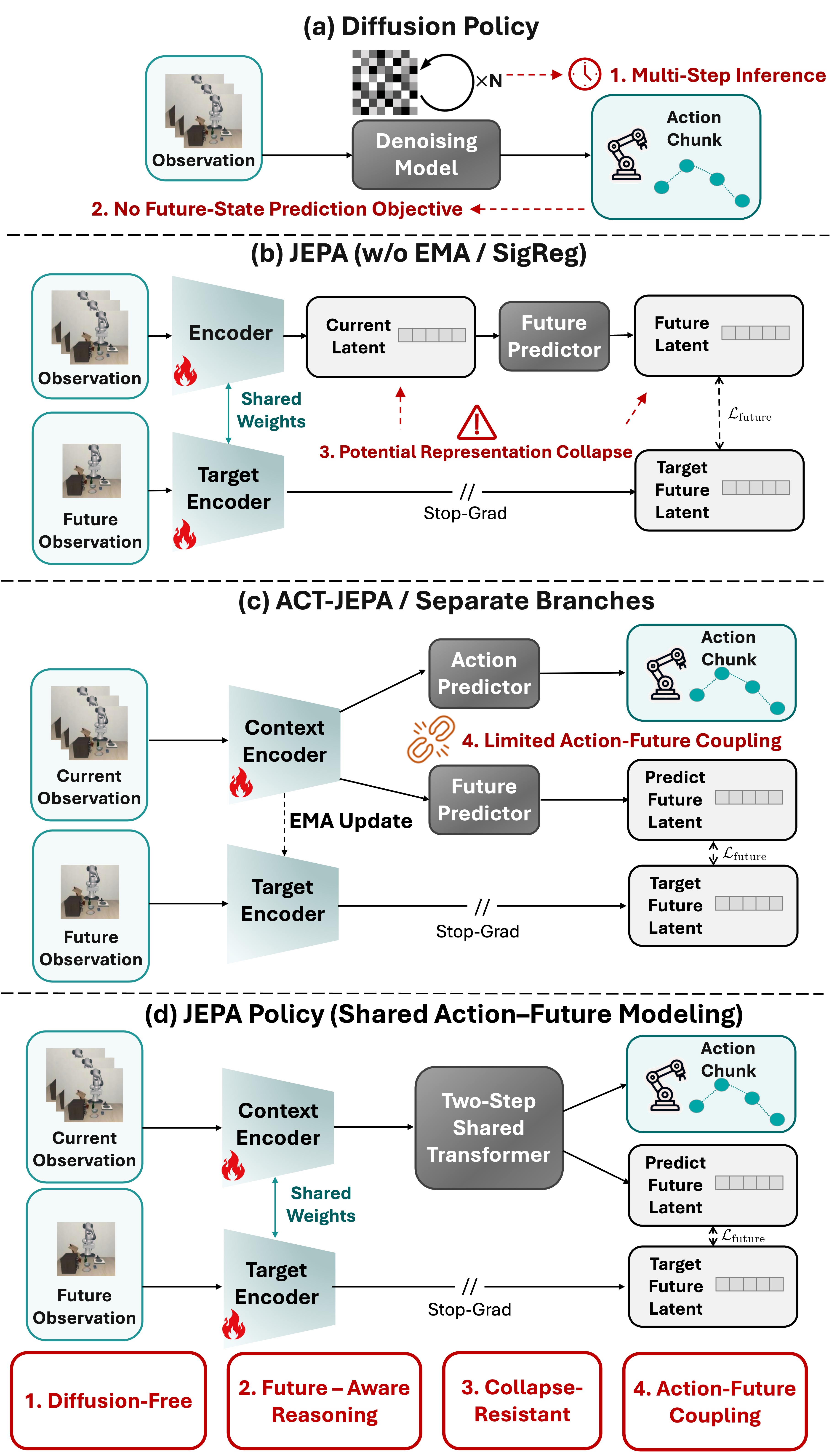}
\caption{
Schematic comparison of design paradigms (architecture only, not a performance comparison). Red numbers mark a limitation of each prior design; the four boxes below show how JEPA Policy addresses them. Flames mark encoders trained by gradient descent. All encoders are trained from scratch because this study uses no pretrained or frozen visual backbone. ``//'' marks the stop-gradient.
}
\label{fig:comparison}
\end{figure}

\section{Introduction}

\IEEEPARstart{R}{obot} imitation learning aims to learn policies capable of completing complex manipulation tasks from expert demonstrations.
Advances in visual encoders, Transformer architectures, and generative action modeling have improved imitation learning on grasping, pushing, insertion, drawer opening, and long-horizon tabletop manipulation.
Typical behavior cloning methods, the paradigm since ALVINN~\cite{alvinn}, formulate policy learning as a supervised problem of predicting the expert action sequence $\mathbf{a}_{t:t+H}$ directly from the current observation $o_t$.
This paradigm is simple and efficient, but its supervision focuses on action labels and does not constrain the resulting future state. It also inherits the compounding error caused by the distribution shift of open-loop supervised imitation~\cite{dagger}.

In robotic manipulation, however, expert actions are not isolated labels.
An action sequence is useful because it advances the current state toward a task-relevant outcome. Examples include pushing an object to a target, approaching a stable grasp, or establishing tool contact.
An expert demonstration segment therefore supplies two aligned labels: the expert action and the future state observed after that action.
Standard behavior cloning typically uses only the action label.

Using these future observations for supervision, however, raises a modeling question.
The same scene may precede a leftward push, a rightward push, or a grasp, each leading to a different outcome.
Regressing these futures from the observation alone can average incompatible targets.
Such ambiguity makes explicit modeling of multiple future modes a natural consideration.
Yet predicting an action's task-relevant consequence is a different objective from generating diverse, visually detailed future videos.

Diffusion Policy~\cite{diffusion_policy} learns an action distribution through iterative denoising, but repeated network calls add inference latency.
The study introducing the \emph{minimal iterative policy} (MIP)~\cite{much_ado_noising} questions whether distribution modeling explains the success of generative control.
On the evaluated behavior-cloning benchmarks, modeling multimodal action distributions did not explain the advantage of flow-based policies.
Instead, supervised iterative computation and suitable training noise were the key ingredients.
Its deterministic two-step MIP matched flow-based policies without fitting the full action distribution.
These findings motivate a further question: can joint prediction of actions and paired future representations improve control without explicitly modeling the multimodal distribution of either output?

Our setting offers two reasons to explore this possibility further.
First, each demonstration pairs an action sequence with the future observation recorded along the same trajectory.
This pairing provides supervision for learning which outcome accompanies which action, rather than treating the two targets as unrelated.
Second, predicting a future representation rather than a video avoids the requirement to reconstruct low-level visual detail.
We therefore extend MIP's coarse-to-refined process to jointly predict actions and their paired future representations.
The aim is to improve control through paired future supervision while retaining two-pass, diffusion-free inference for both outputs.

We propose \textbf{JEPA Policy}, a diffusion-free framework for joint action--future observation representation prediction (Fig.~\ref{fig:comparison}).
Fig.~\ref{fig:comparison} places the design against three reference points. Diffusion Policy~\cite{diffusion_policy}~(a) denoises iteratively, so one decision costs $N$ network calls. Its objective does not constrain the future reached by the predicted action. A plain latent-future JEPA~\cite{ijepa}~(b) has no action supervision or standard anti-collapse mechanism, such as an EMA target encoder or variance--covariance regularization. Its representation can therefore collapse. ACT-JEPA~\cite{act_jepa}~(c) shares a context encoder but uses separate predictor and decoder branches behind an EMA target encoder. The action and future paths consequently remain loosely coupled. JEPA Policy~(d) places action and future representation tokens in one shared two-step Transformer, where they interact in every layer. The same encoder produces the stopped future target, and the model uses no EMA or frozen \emph{target} encoder.

Given the current observation, the model simultaneously predicts the expert action sequence and its paired future observation representation within one shared Transformer.
Action and future-representation tokens share one self-attention stack and remain mutually visible in every layer, allowing the future objective to shape the action-generating representation directly during training.

Both targets are available in the original demonstrations, so the additional supervision requires no new annotation.
Joint prediction encourages the policy to retain information about object position, contact, task progress, and action consequences.
Unlike reconstructing a future image, matching its representation does not require reproducing texture, illumination, or background appearance.

The future prediction branch of JEPA Policy also provides a deployment-time interface.
At inference, the policy outputs the action sequence through the two-step predictor and simultaneously produces a prediction of the future observation representation.
After the action is executed, the actual future observation can be mapped into representation space by the same encoder and compared with the previously predicted future representation.
This future-consistency error serves as a rollout-level diagnostic signal for analyzing whether the policy is drifting away from the expert demonstration manifold, entering high-risk states, or requiring human intervention.
We treat this signal as a behavior-consistency diagnostic, not as a complete uncertainty estimate.

This setting differs from purely self-supervised JEPA-style latent prediction.
In pure self-supervised representation prediction, without additional anti-collapse mechanisms the model may map different inputs to similar or identical representations, forming a trivial solution.
Many self-supervised JEPA or Siamese representation learning methods therefore require an EMA target encoder, a frozen target encoder, architectural asymmetry, or explicit variance--covariance regularization.
In contrast, the future observation representation prediction of JEPA Policy is always optimized jointly with expert action prediction.
If the shared representation collapses, object positions, contact states, and control requirements under different observations become indistinguishable, directly increasing the action prediction error.
Action supervision thus provides a task-driven constraint against collapse and encourages the model to retain information needed for control. This design requires no EMA target encoder, frozen encoder, or explicit variance--covariance regularizer. Section~\ref{subsec:collapse} evaluates the resulting representations.
The visual encoder is initialized randomly and trained end to end with the policy rather than imported from a pretrained representation model.
The observed effect therefore cannot be attributed to a frozen visual backbone that already encodes future-predictive structure.
This setting also makes representation stability an optimization question for the proposed objective itself, rather than a property inherited from external pretraining.

The contributions of this paper are fourfold:

\begin{itemize}
    \item We propose \textbf{JEPA Policy}, a future-aware robot imitation learning framework based on paired action--future observation representation supervision.
    Each demonstration jointly supervises an expert action sequence and its observed future representation, preserving their correspondence within the same trajectory.
    Architecturally, both token types occupy one shared Transformer and remain mutually visible in every self-attention layer, so the future objective directly shapes the action-generating representation.

    \item We extend \textbf{MIP-style coarse-to-refined prediction} to the joint generation of actions and future observation representations.
    A two-step diffusion-free predictor refines the action sequence and its paired future representation within one shared network.
    The resulting policy adds future supervision to MIP's action-only formulation without introducing a distribution-fitting objective or a denoising chain.
    Training and deployment remain diffusion-free, and the policy provides a future latent prediction alongside each action chunk.

    \item We use expert action supervision to provide a \textbf{task-driven anti-collapse constraint} for latent-space future observation representation prediction.
    Unlike purely self-supervised JEPA objectives, the shared representation of JEPA Policy must simultaneously support action prediction and future representation prediction; once the representation collapses, the action error immediately increases.
    Our visual encoder is trained end to end from scratch. The method needs neither an EMA target encoder nor a separately frozen encoder, and it uses no explicit variance--covariance regularizer.

    \item We propose a \textbf{deployment-time future-consistency diagnostic}.
    The future observation representation predicted by the policy can be compared with the actual future observation representation after execution, forming a rollout-level future-consistency error usable for failure analysis, risk monitoring, or potential human intervention.
\end{itemize}

\section{Related Work}
\subsection{Imitation Learning for Robotic Manipulation}
\looseness=-1 Behavior cloning trains policies to match expert actions via supervised learning.
Modern visuomotor imitation learning methods typically combine image encoders, Transformers, and action chunking to model temporally continuous action sequences \cite{act,robomimic}.
These methods require only one direct forward pass at inference.
However, their training objectives focus on action matching and do not explicitly require the policy representation to encode the future outcomes produced by the demonstrated actions.

\subsection{Diffusion Policy and Generative Action Modeling}
Diffusion Policy models action generation as an iterative denoising process conditioned on observations \cite{diffusion_policy}.
This approach can represent complex action distributions and performs well across multiple visuomotor control tasks.
Recent work attributes some gains of diffusion and flow-matching policies to noise injection and denoising training, rather than only to multimodal modeling~\cite{much_ado_noising}.
The primary cost is that inference requires multiple network calls.
Diffusion was not the first approach to multimodal action distributions. Implicit behavior cloning fits an energy-based policy~\cite{ibc}, while Behavior Transformers discretize the action space and predict a residual offset~\cite{bet}. Our additional supervision serves a different purpose. It links each demonstrated action to the future that followed it instead of modeling the distribution of possible actions.

\looseness=1 Few- and one-step alternatives already reduce sampling cost through consistency or flow reformulations~\cite{consistency_models,flow_matching,consistency_policy,shortcut_models}, while DP3~\cite{dp3} reduces observation cost.
We do not benchmark against them and make no relative speed claim.
Our predictor is not a new accelerator; it adds paired future supervision within an existing two-pass \mbox{inference budget}.

\subsection{Predictive Representation Learning and JEPA}
\label{subsec:jepa_related}
Joint-embedding predictive architectures learn by predicting the latent representation of a target view or future observation rather than reconstructing pixels \cite{ijepa,lecun2022path}.
This avoids fully modeling low-level image details but requires handling representation collapse.
Existing methods commonly use architectural asymmetry, stop-gradient, momentum encoders, or variance--covariance regularization \cite{byol,vicreg,barlow}.
Our setting differs: the predictive representation is embedded inside an imitation learning policy, and the shared representation must support accurate expert action prediction.
Most related to this paper is ACT-JEPA \cite{act_jepa}, which also brings JEPA into imitation learning by jointly predicting action sequences and latent representations of future observations.
Its architecture, however, routes a shared context encoder to separate observation-prediction and action-decoding branches. These branches do not interact within one attention stack, and the method relies on an EMA target encoder to prevent collapse.
JEPA Policy instead uses one mutually visible stack, a stopped target as in SimSiam~\cite{simsiam}, and action supervision rather than an EMA or frozen target encoder.
Because non-contrastive objectives can fail through partial rank collapse~\cite{dim_collapse,ssl_dynamics}, Sec.~\ref{subsec:collapse} audits the representation spectrum.
GR-1~\cite{gr1} and Seer~\cite{seer} also predict future visuals and actions in one Transformer.
JEPA Policy instead uses a paired latent target, no video pretraining, two fixed feed-forward steps, and a shared-stack topology tested in Sec.~\ref{subsec:future_ablation}.
This distinction separates the contribution from transferring a pretrained world representation into a policy.
Our experiments ask whether paired future supervision can improve action learning when the encoder and policy are trained together from scratch on the task demonstrations.
They do not test the additional benefit that large-scale video or vision-language pretraining might provide.

A parallel line scales imitation with multi-task or pretrained VLA backbones, including RT-1~\cite{rt1}, RT-2~\cite{rt2}, OpenVLA~\cite{openvla}, Octo~\cite{octo}, Open X-Embodiment~\cite{openx}, and $\pi_0$~\cite{pi0}.
Concurrent VLA-JEPA~\cite{vlajepa} predicts leakage-free future latents with a pretrained VLM and flow matching, while JEPA-VLA~\cite{jepavla} injects video-predictive embeddings into existing VLAs.
JEPA Policy instead trains on each task's demonstrations, shares action and future tokens in one stack, uses two MIP steps, and relies on action loss against collapse.

\subsection{World Models and Future Prediction}
World and forward models predict action-conditioned futures for planning or model-predictive control~\cite{worldmodels}.
V-JEPA~2-AC~\cite{vjepa2} uses a frozen video-pretrained encoder and latent predictor for image-goal MPC.
CURL~\cite{curl}, SPR~\cite{spr}, DreamerV3~\cite{dreamerv3}, and TD-MPC2~\cite{tdmpc2} likewise learn representations or dynamics for rollout.
JEPA Policy learns no separate dynamics model, performs no planning rollout, and reconstructs no future images.
It consumes one demonstrated future latent paired with the supervised action as a policy-training signal.

\subsection{Runtime Failure Monitoring for Learned Policies}
Runtime failure detection for learned policies draws on a broad literature in predictive uncertainty. Representative approaches include softmax confidence for out-of-distribution inputs~\cite{msp_ood}, deep ensembles~\cite{deep_ensembles}, and conformal prediction. The latter provides distribution-free safety margins and has been applied to planning among dynamic agents~\cite{conformal_planning}.
Sentinel~\cite{sentinel} combines temporal action consistency with a vision--language model that evaluates task progress. FAIL-Detect~\cite{fail_detect} detects failures in imitation policies without requiring failure data.
Our diagnostic has a narrower scope. It is a by-product of the training objective rather than a purpose-built monitor. Because the policy already predicts a future representation, comparison with the realized representation requires one encoder pass and no additional model.
We therefore present it as a per-task-calibrated risk signal, not as a competitor to dedicated monitors. A direct comparison with such monitors remains future work.

\begin{figure*}[!t]
\centering
\includegraphics[width=\textwidth]{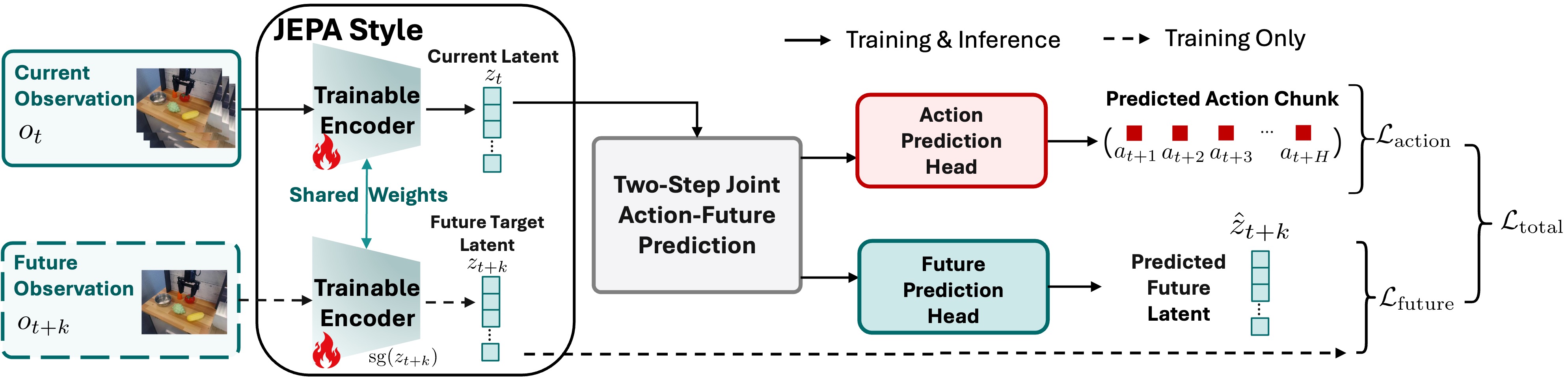}
\caption{
Overall architecture of JEPA Policy.
A shared encoder processes both observations. $\mathbf{z}_t$ and $\mathbf{z}_{t+k}$ are the outputs of its image path at the two times. The conditioning vector $\mathbf{c}_t$ fuses $\mathbf{z}_t$ with low-dimensional inputs (Sec.~\ref{subsec:architecture}); the fusion is omitted from the diagram. Action tokens $\mathbf{q}_a$ and future-representation tokens $\mathbf{q}_z$ enter one Transformer $T_\theta$ and interact in every self-attention layer. Separate heads produce the action sequence $\hat{\mathbf{a}}_{t:t+H}$ and future observation representation $\hat{\mathbf{z}}_{t+k}$. The same encoder produces the stopped target $\mathbf{z}_{t+k}=E_{\mathrm{rgb}}(o_{t+k})$. Shared processing allows the future-prediction objective to update the representation used by the action head.
}
\label{fig:architecture}
\end{figure*}

\section{Method}
\label{sec:method}

This section specifies JEPA Policy's paired objective, shared stack, two-step training rule, and deployment diagnostic.

\subsection{Problem Formulation}
\label{subsec:problem}

Given an expert demonstration dataset
\begin{equation}
    \mathcal{D} =
    \{(o_t, \mathbf{a}_{t:t+H}, o_{t+k})\},
\end{equation}
where $o_t$ denotes the current observation or observation history window, $\mathbf{a}_{t:t+H}$ denotes the expert action sequence over a prediction horizon of $H$, and $o_{t+k}$ denotes the future observation at horizon $k$.
We use $\mathbf{a}_{t:t+H}$ as compact notation for the next $H$ actions, $(\mathbf{a}_{t+1},\ldots,\mathbf{a}_{t+H})$.
Both $H$ and $k$ are counted in environment control steps; their wall-clock meaning is quantified in Sec.~\ref{subsec:exp_setup}.
The horizon $k$ sets how far the target observation lies in the future. If $k$ is too small, $\mathbf{z}_{t+k}$ closely resembles the current observation and provides little new information. If $k$ is too large, the current action chunk has little influence on the target and the pairing becomes weak.
We therefore set $k$ on the same order of magnitude as the action horizon $H$ and determine its concrete value by ablation.
Conventional direct policy learning formulates imitation learning as a supervised mapping from the current observation to expert actions:
\begin{equation}
    \pi_\theta(o_t) \rightarrow \hat{\mathbf{a}}_{t:t+H}.
\end{equation}
This formulation requires the policy only to predict expert actions, without explicitly constraining the future outcomes that action sequence induces in the demonstration trajectory.

JEPA Policy learns the following joint mapping:
\begin{equation}
    \pi_\theta(o_t) \rightarrow
    \{\hat{\mathbf{a}}_{t:t+H}, \hat{\mathbf{z}}_{t+k}\},
\end{equation}
where $\hat{\mathbf{a}}_{t:t+H}$ is the predicted action sequence and $\hat{\mathbf{z}}_{t+k}$ is the predicted future observation representation.
Here $\pi_\theta(o_t)$ is a shorthand for the final composite policy.
For a unified notation, we write the network as the coarse-to-refined predictor $F_\theta(o_t, \mathbf{a}_{\mathrm{in}}, \mathbf{z}_{\mathrm{in}}, u, v)$.
Its inputs are the current observation, an action input, a future-representation input, and the interval endpoints $(u,v)$. The action input can be zero or noisy.
The network outputs an action and a future observation representation. $F_\theta$ denotes the shared Transformer policy described in Sec.~\ref{subsec:architecture}; the two names refer to the same network.
The future observation representation target is obtained by passing the actual future observation through the observation encoder:
\begin{equation}
    \mathbf{z}_{t+k} = E_{\mathrm{rgb}}(o_{t+k}),
\end{equation}
where $E_{\mathrm{rgb}}(\cdot)$ is the image path of the same observation encoder $E$. It globally pools and concatenates the per-camera backbone features before they are fused with low-dimensional inputs.
The conditioning representation $\mathbf{c}_t = E(o_t)$ is the output of that fusion layer and therefore also carries proprioception, whereas the future target is image-only and higher-dimensional (1024 against 384 in our configuration).
For the current observation, the same image path gives $\mathbf{z}_t = E_{\mathrm{rgb}}(o_t)$. The conditioning representation is $\mathbf{c}_t = \mathrm{Fuse}\big([\mathbf{z}_t,\ \mathbf{s}_t]\big)$, where $\mathbf{s}_t$ contains the low-dimensional proprioceptive inputs.
Figure~\ref{fig:architecture} shows $\mathbf{z}_t$ and $\mathbf{z}_{t+k}$ as outputs of the same function at two times and leaves the fusion step implicit.
Both use one encoder and the same image-backbone parameters. Their values differ because they are computed from observations at different times.
Unlike future image reconstruction, JEPA Policy predicts future observation representations only in latent space, avoiding the modeling of low-level visual details such as texture, illumination, and background that are weakly related to robot control decisions.

The shared encoder $E$ is trained end to end, without a frozen or EMA target encoder.
We stop gradients only through the future target $\mathbf{z}_{t+k}$; $\mathcal{L}_{\mathrm{future}}$ still updates $E$ through the current-observation path $\mathbf{c}_t$, as detailed in Sec.~\ref{subsec:joint_loss}.

\subsection{Shared Action--Future Observation Representation Policy}
\label{subsec:architecture}

The overall structure of JEPA Policy is shown in Fig.~\ref{fig:architecture}.
The policy is exactly the $F_\theta(o_t,\mathbf{a}_{\mathrm{in}},\mathbf{z}_{\mathrm{in}},u,v)$ introduced in Sec.~\ref{subsec:problem}, consisting of an observation encoder, $H$ action input tokens, a single future input token, a shared Transformer, an action prediction head, and a future observation representation prediction head.
The current observation is first encoded into a conditioning representation:
\begin{equation}
    \mathbf{c}_t = E(o_t).
\end{equation}
Together with an embedding of the two interval endpoints $(u,v)$ of the current step, $\mathbf{c}_t$ forms the conditioning memory
\begin{equation}
    \mathbf{M} = \mathrm{Enc}\big([\,\mathrm{TimeEmb}(u,v),\ \mathrm{Embed}_o(\mathbf{c}_t)\,]\big),
    \label{eq:memory}
\end{equation}
which the shared Transformer reads by cross-attention.
Separate linear projections embed the action input $\mathbf{a}_{\mathrm{in}}$ and future-representation input $\mathbf{z}_{\mathrm{in}}$. The resulting tokens occupy $H+1$ positions in one sequence. A learned positional embedding $\mathbf{p}$ covers all positions, while $\mathbf{e}_z$ marks the future position:
\begin{align}
    \mathbf{q}_a &= \mathrm{Embed}_a(\mathbf{a}_{\mathrm{in}}) + \mathbf{p}_{1:H}, \\
    \mathbf{q}_z &= \mathrm{Embed}_z(\mathbf{z}_{\mathrm{in}}) + \mathbf{e}_z + \mathbf{p}_{H+1}.
\end{align}
The step parameters are deliberately \emph{not} added to these tokens; they enter only through $\mathbf{M}$ in Eq.~\eqref{eq:memory}.
The shared Transformer then jointly models the two token types:
\begin{equation}
    \mathbf{Y} = T_\theta([\mathbf{q}_a, \mathbf{q}_z], \mathbf{M}),
\end{equation}
where $T_\theta$ denotes the shared policy Transformer (the backbone of $F_\theta$), $[\cdot,\cdot]$ denotes token concatenation, and each of its layers applies unmasked self-attention over the $H+1$ concatenated tokens followed by cross-attention to $\mathbf{M}$.
The output $\mathbf{Y}$ is split into action token representations $\mathbf{Y}_a$ and future token representations $\mathbf{Y}_z$:
\begin{equation}
    \mathbf{Y}_a, \mathbf{Y}_z = \mathrm{split}(\mathbf{Y}).
\end{equation}
The two prediction heads then output the action sequence and the future observation representation, respectively:
\begin{align}
    \hat{\mathbf{a}}_{t:t+H} &= h_a(\mathbf{Y}_a), \\
    \hat{\mathbf{z}}_{t+k} &= h_z(\mathbf{Y}_z),
\end{align}
where $h_a(\cdot)$ denotes the action prediction head and $h_z(\cdot)$ denotes the future observation representation prediction head.
The overall mapping is exactly $F_\theta(o_t,\mathbf{a}_{\mathrm{in}},\mathbf{z}_{\mathrm{in}},u,v)=(\hat{\mathbf{a}}_{t:t+H},\hat{\mathbf{z}}_{t+k})$.
The two steps use interval endpoints $(u,v)=(0,\tau)$ and $(\tau,1)$. In the first step, $\mathbf{a}_{\mathrm{in}}=\mathbf{0}$ and $\mathbf{z}_{\mathrm{in}}=\mathbf{0}$.
The tokens then reduce to learned queries formed by the input-projection biases, $\mathbf{p}$, and $\mathbf{e}_z$. This step regresses a coarse prediction from the current observation.
The second step receives noised expert targets during training and the first-step predictions during inference. It then refines both outputs.

In this structure, future observation representation prediction is not an isolated auxiliary branch appended after the policy.
Shared attention allows future prediction to shape the policy representation used for action generation.
The action and future tokens occupy one unmasked self-attention stack and can interact in every layer. The action representation $\mathbf{Y}_a$ therefore contains information from the future token during the forward pass.
The future loss also backpropagates through the same attention layers that produce the action representation.
Separate-branch designs such as ACT-JEPA~\cite{act_jepa} behave differently. Their future loss can update a shared encoder, but future predictions do not enter the computation of the action tokens.
The shared stack thus creates two routes from future prediction to action generation.
The forward route lets action tokens attend to the future token during the policy computation.
The backward route lets the future loss update the same intermediate parameters used by the action head.
A separate future branch may retain the second route only through an upstream encoder, while remaining absent from the computation that produces the action tokens.
This distinction motivates both the shared-versus-separate comparison and the route-cut controls in the experiments.
Section~\ref{subsec:future_ablation} tests whether this difference in topology accounts for the performance gain.

\subsection{Joint Action--Future Representation Objective}
\label{subsec:joint_loss}

The action prediction loss is defined as
\begin{equation}
    \mathcal{L}_{\mathrm{act}} =
    \big\langle\,\big(\hat{\mathbf{a}}_{t:t+H} - \mathbf{a}_{t:t+H}\big)^2\,\big\rangle,
\end{equation}
where $\langle\cdot\rangle$ denotes the mean over the action dimension, the horizon $H$ and the batch, so that the loss scale is independent of $H$ and of the action dimensionality.
The future observation representation prediction loss aligns the predicted future representation with the target representation of the actual future observation in the demonstration:
\begin{equation}
    \mathcal{L}_{\mathrm{future}} =
    d_{\mathrm{NMSE}}\left(\hat{\mathbf{z}}_{t+k}, \mathrm{sg}(\mathbf{z}_{t+k})\right),
\end{equation}
where $d_{\mathrm{NMSE}}(\hat{\mathbf{z}},\mathbf{z})=\| (\hat{\mathbf{z}}-\mathbf{z})/\max(\|\mathbf{z}\|_{\mathrm{RMS}},10^{-6})\|_2^2/d_z$, and $d_z$ is the latent dimension. We then average over the batch. This target-normalized MSE separates the latent-space scale from the action-space scale and simplifies loss balancing (Sec.~\ref{subsec:exp_setup}).
$\mathrm{sg}(\cdot)$ denotes a stop-gradient that we \textbf{always} apply to the future target representation $\mathbf{z}_{t+k}=E_{\mathrm{rgb}}(o_{t+k})$.
The current and future observations use the same encoder parameters $E$. The future loss still updates this encoder because $\hat{\mathbf{z}}_{t+k}$ depends on $\mathbf{c}_t=E(o_t)$. Gradients therefore reach $E$ through the current-observation path.
The stop-gradient cuts only the future-observation target path $\mathbf{z}_{t+k}=E_{\mathrm{rgb}}(o_{t+k})$, preventing direct target-side co-adaptation but not, by itself, excluding a constant representation.
The future loss still shapes the encoder through the current-observation path, while the action loss $\mathcal{L}_{\mathrm{act}}$ supplies the task-driven constraint tested in Sec.~\ref{subsec:collapse}.
In implementation, both the action loss and the future observation representation loss are averaged over the batch and output dimensions.

The overall training objective is
\begin{equation}
    \mathcal{L} =
    \lambda_a \mathcal{L}_{\mathrm{act}} +
    \lambda_z \mathcal{L}_{\mathrm{future}},
    \label{eq:joint_loss}
\end{equation}
The coefficients $\lambda_a$ and $\lambda_z$ weight these two terms.

Two settings of $\lambda_z$ are implemented. Fixed mode uses a constant and appears only where stated. Unless otherwise stated, the main experiments use \emph{adaptive ratio} mode, which recomputes $\lambda_z$ once per minibatch:
\begin{equation}
    \lambda_z = \mathrm{clip}\!\left(
      \rho\,\frac{\mathrm{sg}(\lambda_a\mathcal{L}_{\mathrm{act}})}
                 {\mathrm{sg}(\mathcal{L}_{\mathrm{future}})+\varepsilon},
      \;\lambda_{\min},\;\lambda_{\max}\right),
    \label{eq:ratio_mode}
\end{equation}
For the two-step objective, the losses in Eq.~\eqref{eq:ratio_mode} sum the unweighted, interval-scaled components from both steps; one $\lambda_z$ is shared by both. Stop-gradient applies only inside Eq.~\eqref{eq:ratio_mode}, so $\lambda_z$ carries no gradient; both terms in Eq.~\eqref{eq:joint_loss} retain their normal gradients. We set $\varepsilon=10^{-8}$ and, in adaptive-ratio experiments, $\lambda_a=1$, default $\rho=0.10$, and $[\lambda_{\min},\lambda_{\max}]=[10^{-6},0.1]$. We ablate $\rho\in\{0.05,0.10,0.20\}$ in Sec.~\ref{subsec:ratio_horizon}. Without clipping, $(\lambda_z\mathcal{L}_{\mathrm{future}})/(\lambda_a\mathcal{L}_{\mathrm{act}})\approx\rho$. The upper and lower bounds can move this ratio below and above $\rho$, respectively. Thus, $\rho$ is neither an exact constraint nor the future term's share of total loss.

Through Eq.~\eqref{eq:joint_loss}, action prediction and future observation representation prediction are jointly optimized in the same policy network, introducing paired action--future supervision for diffusion-free imitation learning.

\subsection{MIP-Style Two-Step Prediction Training}
\label{subsec:mip}

To avoid the inference overhead of the long-chain multi-step denoising in diffusion policies, JEPA Policy adopts a MIP-style two-step prediction process.
MIP uses \textbf{supervised iterative computation} and \textbf{stochasticity injection}. Both prediction steps are supervised, and the second-step input contains noise.
MIP does not learn a continuous velocity field or require multi-step integration. Our MIP-style implementation directly regresses the target at both steps and scales the two residuals by their interval lengths. The target can be an action or a future observation representation.
We apply this process simultaneously to the expert action sequence and the future observation representation.
The two-step interpolation parameter $\tau$ is a fixed hyperparameter ($\tau\in(0,1)$, default $\tau=0.9$) that takes the same value during training and inference, and it is unrelated to the time-step index $t$ used in subscripts such as $o_t$ and $\mathbf{a}_{t:t+H}$.

The first step takes zero vectors as input at time index $0$ and directly produces coarse actions and coarse future observation representations from the current observation:
\begin{equation}
    (\hat{\mathbf{a}}^{(0)}_{t:t+H}, \hat{\mathbf{z}}^{(0)}_{t+k})
    =
    F_\theta\!\left(o_t,\;\mathbf{0},\;\mathbf{0},\;0,\;\tau\right).
\end{equation}
The first-step loss is
\begin{equation}
\begin{aligned}
    \mathcal{L}^{(0)}
    &= \lambda_a\Big\langle\Big(\tfrac{\hat{\mathbf{a}}^{(0)}_{t:t+H}-\mathbf{a}_{t:t+H}}{\tau}\Big)^2\Big\rangle \\
    &\quad + \frac{\lambda_z}{\tau^2}
    d_{\mathrm{NMSE}}\!\left(\hat{\mathbf{z}}^{(0)}_{t+k},\mathrm{sg}(\mathbf{z}_{t+k})\right).
\end{aligned}
\end{equation}

The second step constructs noisy inputs near the expert target.
Given Gaussian noise $\boldsymbol{\epsilon}_a,\boldsymbol{\epsilon}_z\sim\mathcal{N}(\mathbf{0},\mathbf{I})$, define
\begin{align}
    \tilde{\mathbf{a}}_{t:t+H} &= \mathbf{a}_{t:t+H} + (1-\tau)\boldsymbol{\epsilon}_a, \\
    \tilde{\mathbf{z}}_{t+k} &= \mathrm{sg}(\mathbf{z}_{t+k}) + (1-\tau)\boldsymbol{\epsilon}_z.
\end{align}
The model then refines both inputs back to the expert action and the true future observation representation at time index $\tau$:
\begin{equation}
    (\hat{\mathbf{a}}^{(1)}_{t:t+H}, \hat{\mathbf{z}}^{(1)}_{t+k})
    =
    F_\theta\!\left(o_t,\;\tilde{\mathbf{a}}_{t:t+H},\;\tilde{\mathbf{z}}_{t+k},\;\tau,\;1\right).
\end{equation}
The second-step loss is
\begin{equation}
\begin{aligned}
    \mathcal{L}^{(1)}
    &= \lambda_a\Big\langle\Big(\tfrac{\hat{\mathbf{a}}^{(1)}_{t:t+H}-\mathbf{a}_{t:t+H}}{1-\tau}\Big)^2\Big\rangle \\
    &\quad + \frac{\lambda_z}{(1-\tau)^2}
    d_{\mathrm{NMSE}}\!\left(\hat{\mathbf{z}}^{(1)}_{t+k},\mathrm{sg}(\mathbf{z}_{t+k})\right).
\end{aligned}
\end{equation}
\begin{figure}[t]
\centering
\includegraphics[width=0.76\columnwidth]{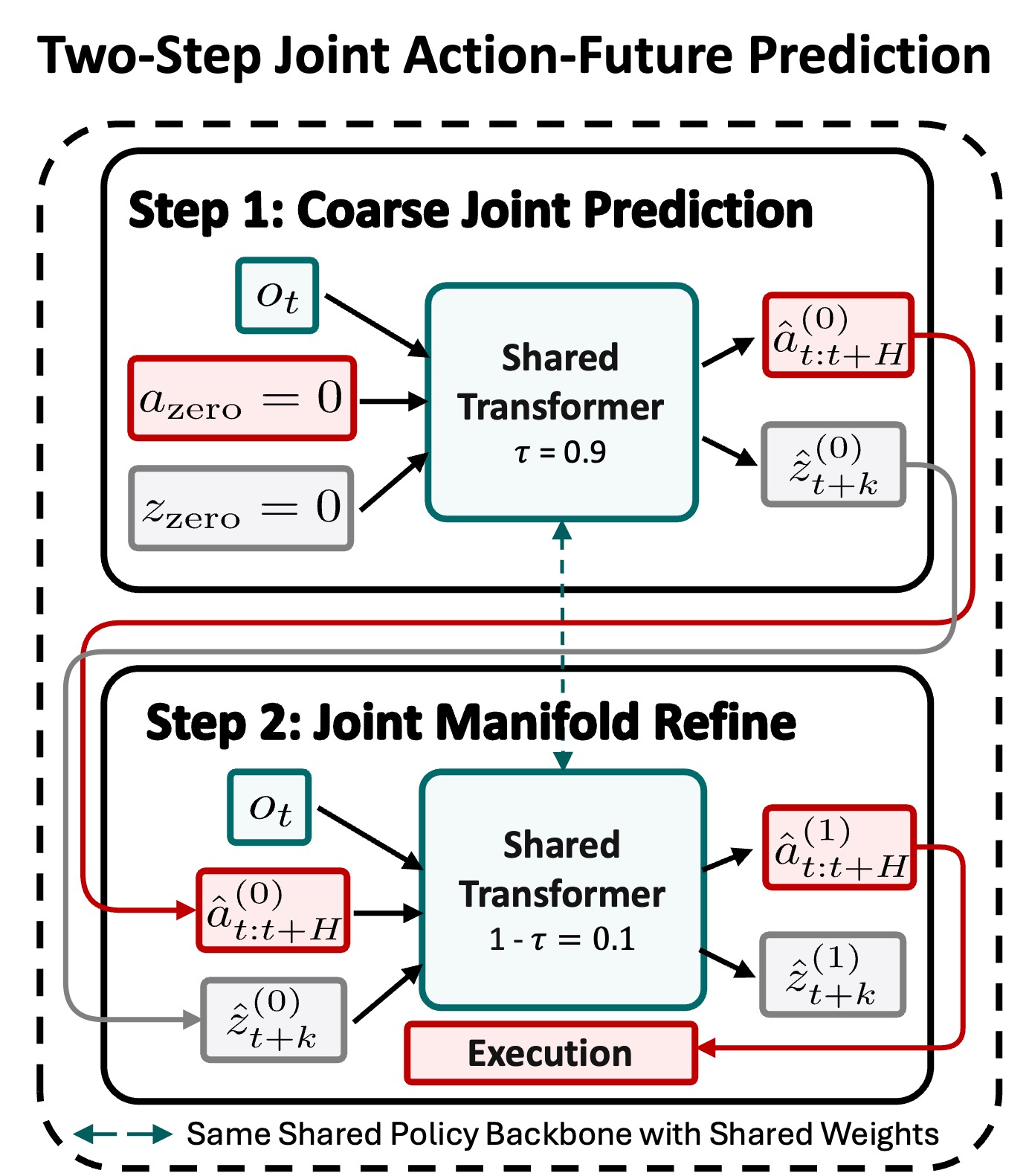}
\caption{
Coarse-to-refined MIP-style training for joint action and future observation representation prediction.
The first step maps a zero input (time $0$) to a coarse prediction, and the second refines it (time $\tau$). The diagram shows inference, where the second step receives the first-step output. During training, it instead receives the noised expert target from Sec.~\ref{subsec:mip}. Both steps are supervised, with residuals scaled by $\tfrac{1}{\tau}$ and $\tfrac{1}{1-\tau}$. The method uses no diffusion-style iterative sampling.
}
\label{fig:mip_training}
\end{figure}

The two-step residuals are divided by $\tau$ and $1-\tau$ to place them on a common scale.
The second-step input differs from the expert target by $(1-\tau)\boldsymbol{\epsilon}$. Division by $1-\tau$ therefore yields a unit-scale correction.
The first-step input is zero rather than a perturbed target, so the same derivation does not apply to that step.
Its division by $\tau$ is the matching convention for an interval of length $\tau$ and should not be interpreted as correcting injected noise.
The two steps are decoupled during training: the input of the second step is constructed from the ground-truth target and noise, not from the output of the first step.
The two supervised prediction steps (iterative computation) and the input noise of the second step (stochasticity injection) together constitute the two effective ingredients of MIP.
The final joint MIP objective is
\begin{equation}
    \mathcal{L}_{\mathrm{MIP}}
    =
    \mathbb{E}_{\mathcal{D},\boldsymbol{\epsilon}_a,\boldsymbol{\epsilon}_z}
    \big[\mathcal{L}^{(0)} + \mathcal{L}^{(1)}\big].
    \label{eq:mip_loss}
\end{equation}
The expectation is over training tuples and both noise variables.
We use the simplified second-step input $\mathbf{a}_{t:t+H}+(1-\tau)\boldsymbol{\epsilon}_a$ rather than the original $\tau\,\mathbf{a}_{t:t+H}+(1-\tau)\boldsymbol{\epsilon}_a$; the two coincide as $\tau$ approaches $1$.

Fig.~\ref{fig:mip_training} illustrates the two-step MIP-style joint training procedure.
At inference, JEPA Policy first evaluates the zero-input coarse step above with the same fixed $\tau$, then supplies both coarse outputs to a noise-free refinement:
\begin{equation}
    (\hat{\mathbf{a}}^{(1)}_{t:t+H}, \hat{\mathbf{z}}^{(1)}_{t+k})
    =
    F_\theta\!\left(o_t,\;\hat{\mathbf{a}}^{(0)}_{t:t+H},\;\hat{\mathbf{z}}^{(0)}_{t+k},\;\tau,\;1\right).
\end{equation}
The refined action $\hat{\mathbf{a}}^{(1)}_{t:t+H}$ controls the robot.
In the original MIP formulation, the second-step input is the first-step output scaled by $\tau$, i.e., $\tau\,\hat{\mathbf{a}}^{(0)}_{t:t+H}$; we adopt the simplified form above, and the two coincide as $\tau$ approaches $1$.

\subsection{Future Observation Representation Prediction under Paired Action Supervision}
\label{subsec:diffusion_free_future}

Predicting a future observation representation from the current observation alone is under-constrained.
Similar observations may precede different futures, so plain regression may average targets or encode control-irrelevant variation.

JEPA Policy instead uses the demonstration tuple of Sec.~\ref{subsec:problem}: the action target and $\mathbf{z}_{t+k}$ from the same segment jointly supervise the two outputs.
At the supervision level, each segment supplies both labels; at the representation level, the future loss reaches the action representation through the shared attention stack.
Each training sample therefore preserves the correspondence between one demonstrated action and its observed outcome. The predictor remains deterministic and can still average across ambiguous repeated observations.
This construction neither assumes a one-to-one action--future mapping nor recovers every possible future mode. It uses paired evidence more fully while retaining the limitations of deterministic regression. The future branch shapes the policy during training and supplies the diagnostic defined next for deployment-time use.

\subsection{Deployment-Time Future-Consistency Diagnostics}
\label{subsec:future_consistency}

At inference, JEPA Policy outputs the refined control action sequence together with the refined future observation representation $\hat{\mathbf{z}}^{(1)}_{t+k}$.
After the policy executes the action, the actual future observation $o_{t+k}^{\mathrm{real}}$ becomes available, and the same observation encoder produces its representation:
\begin{equation}
    \mathbf{z}_{t+k}^{\mathrm{real}} = E_{\mathrm{rgb}}(o_{t+k}^{\mathrm{real}}).
\end{equation}
We accordingly define the rollout-level future-consistency error:
\begin{equation}
    e_{\mathrm{fc}}(t,k)
    = d_{\mathrm{NMSE}}\left(\hat{\mathbf{z}}^{(1)}_{t+k},
    \mathrm{sg}\left(\mathbf{z}_{t+k}^{\mathrm{real}}\right)\right).
\end{equation}
When policy behavior remains close to the demonstrations, the predicted representation should agree with the observed post-execution representation.
A large $e_{\mathrm{fc}}$ may instead indicate an unexpected behavior, an abnormal contact outcome, or a visual state outside the training distribution.
We treat this quantity as a diagnostic signal for failure analysis and human-intervention triggering, not as a strict probabilistic uncertainty estimate.
In practical deployment, the alarm threshold can be chosen using a sliding-window average $\bar e_{\mathrm{fc}}$, task-phase-conditioned thresholds, or ROC curves computed from successful/failed validation rollouts.
The signal becomes available only after the corresponding future observation has been reached, so it diagnoses the preceding prediction rather than preventing that action from executing.
Its operational use is to inform the next control decision, request intervention, or annotate a rollout for later analysis.
Section~\ref{subsec:diag_results} therefore evaluates ranking and calibration separately instead of treating the raw error as an instantaneous safety guarantee.

\section{Experiments}
\label{sec:experiments}

\looseness=-1 We evaluate JEPA Policy on nine simulated manipulation tasks from three benchmark suites, using three training seeds per configuration.
The evaluation also includes a pre-specified collapse audit on seven tasks over 84 checkpoints (Sec.~\ref{subsec:collapse}) and a 5{,}200-episode diagnostic audit on the original eight-task set (Sec.~\ref{subsec:diag_results}).
Sec.~\ref{subsec:real_robot} repeats the comparison on physical hardware.

\subsection{Setup}
\label{subsec:exp_setup}

\textbf{Tasks.}
We use tasks from LIBERO~\cite{libero}, robomimic~\cite{robomimic}, and MimicGen~\cite{mimicgen}. These suites are built on robosuite/MuJoCo~\cite{robosuite,mujoco}.
LIBERO provides MokaMoka and MugMug, with $50$ demonstrations and $20{,}794/12{,}909$ transitions. Robomimic provides Square PH, Tool Hang PH, and four-camera Transport PH, with $200$ demonstrations and $30{,}154/95{,}962/93{,}752$ transitions. MimicGen provides Coffee Preparation D1, Kitchen D1, Three Piece Assembly D1, and Hammer Cleanup D1, each with $1{,}000$ demonstrations and $286{,}847$ to $687{,}674$ transitions.
The four-camera Transport dataset (four $84{\times}84$ RGB views) was generated for this work and passed an independent structural audit; a two-camera variant exists for one seed and is excluded from the main comparison.

\textbf{Protocol.}
The primary MIP, Diffusion Policy, and JEPA Policy runs train for 300k gradient steps. We use batch size 256, learning rate $10^{-4}$, and EMA rate 0.995. Each run is evaluated every 10k steps under a fixed evaluation seed, giving 30 rollout evaluations.
We report the maximum success rate over the 30 evaluations (best-checkpoint protocol), averaged over seeds $\{41,42,43\}$.
The same checkpoint rule and rollout budget are applied to every primary arm, so selection optimism affects their absolute levels in the same direction.
Table~\ref{tab:protocol} separately reports three selection-free summaries to test whether the difference between JEPA Policy and MIP depends on this rule.
The controlled topology and route-cut comparisons on Transport use an aligned 10k--230k window (23 evaluations); all reported maxima for those comparisons occur within this window.
The reported success rates have finite rollout resolution; differences corresponding to only one or two episode outcomes should not be over-interpreted.

\textbf{Timing.}
All simulated environments run at 20\,Hz control.
The policy predicts $H{=}10$ actions on robomimic and MimicGen and $H{=}16$ on LIBERO. In both cases, it executes the first eight actions before replanning. Actions run at 20\,Hz, and the policy produces a new chunk at 2.5\,Hz.
The future horizon $k$ counts environment transitions after the newest observation frame: the default $k{=}4$ target lies $0.2$\,s ahead ($k{=}2/6/8$: $0.1/0.3/0.4$\,s), i.e., within the $0.4$\,s duration of the currently executed action chunk itself.

\textbf{Implementation details.}
The observation encoder $E$ is a ResNet-18~\cite{resnet} image encoder trained jointly with the policy by default; a frozen DINOv2~\cite{dinov2} ViT-S/14 encoder (CLS token) is supported as an alternative backbone.
The future target is a single pooled global embedding rather than a sequence of patch tokens. It uses only the image channels at the input of the encoder's fusion layer, excluding low-dimensional inputs such as proprioception. We apply stop-gradient to this target.
To preserve the spatial correspondence between current and future observations under image augmentation, the two share the same random crop window during training (temporal-consistent crop).
The future loss uses target-RMS-normalized MSE. We set the interpolation parameter to $\tau=0.9$ and the default future horizon to $k=4$. Inference uses the EMA weights.
This exponential moving average is Polyak averaging of the trained policy for evaluation and is applied to all methods, including the baseline. It is not a BYOL-style EMA \emph{target} encoder. JEPA Policy produces its future target with the live shared encoder and applies stop-gradient (Sec.~\ref{subsec:architecture}).
In the main experiments, the action loss weight is $\lambda_a=1$, and the future loss weight follows Eq.~\eqref{eq:ratio_mode} with target ratio $\rho=0.10$ and clip bounds $[10^{-6},0.1]$. We ablate $\rho\in\{0.05,0.10,0.20\}$.
Peak device memory remains below $6$\,GiB per training run. On Tool Hang, JEPA Policy allocates $5.28$\,GiB and reserves $5.71$\,GiB, compared with $5.02$\,GiB allocated by MIP. Peak inference activations are $362/355$\,MiB for JEPA Policy/MIP (Table~\ref{tab:latency}).

\subsection{Compared Methods}
\label{subsec:baselines}

Table~\ref{tab:main} contrasts the five arms in few-step inference, future-representation supervision, shared attention, and rollout diagnostics. Implementation details follow.

\begin{itemize}
\item \textbf{MIP (action-only)}: the minimal iterative policy of \cite{much_ado_noising}, run from the released implementation that this work extends. Setting $\lambda_z=0$ and removing the future tokens recovers MIP exactly. This arm is therefore both a published baseline and an exact ablation of JEPA Policy. The architecture, optimizer, data pipeline, and evaluation harness are shared by construction.
\item \textbf{Diffusion Policy}~\cite{diffusion_policy}: the diffusion \emph{transformer} policy with a 10-step action horizon. This choice matches the JEPA Policy action-chunk length on the robomimic and MimicGen tasks. The aligned protocol uses the same tasks and seeds, 300k gradient steps, batch size 256, and 30 rollout evaluations under the same task-specific budget. We apply the same best-checkpoint rule to its \texttt{test/mean\_score}. The sampler uses the 100-step DDPM configuration on which it was trained. For each task, we report the better of the absolute and delta action parameterizations because the absolute-action variant failed to learn three robomimic tasks.
    On the two LIBERO tasks, JEPA Policy uses a 16-step horizon, so the horizon match is exact only outside LIBERO. Diffusion Policy is weakest on Tool Hang ($46.7$), which robomimic labels as a high-precision task. On this task, accumulated pose error can affect the final insertion.
\item \textbf{Dual-independent (ACT-JEPA-style topology)}: the shared conditioning path contains the observation embedding, condition encoder, future-memory adapter, and two interval embeddings ($1.40$\,M parameters). It produces a fixed observation memory for two parameter-independent decoder stacks. The action stack has $18.93$\,M parameters and the future stack has $4.75$\,M; both use pre-norm self-attention, fixed-memory cross-attention, and an FFN. No path connects the stacks, so each loss produces zero gradient on the other stack. Both losses still train the shared conditioning path because it has no stop-gradient. This configuration isolates the separate-branch property of ACT-JEPA. To avoid confounding topology with other design choices, it retains our visual target, stopped-target action supervision, and training recipe instead of reproducing the full ACT-JEPA method.
\item \textbf{Dual-cross}: identical to Dual-independent plus a bidirectional cross-attention block at every layer ($5.35$\,M). Each layer applies self-attention on both branches, exchanges information through the cross block, then applies the memory-attention and FFN steps. The two branches are therefore mutually visible, but through an explicit cross module rather than by sharing one attention stack.
\item For reference, JEPA Policy itself passes the concatenation of action tokens $\mathbf{q}_a$ and future tokens $\mathbf{q}_z$ through a single stack ($20.27$\,M) under full bidirectional self-attention with no causal, token-type or memory mask. All three variants share the encoder ($22.90$\,M), the losses, the MIP two-step schedule, the stop-gradient target and $\rho=0.10$; totals are $43.96$\,M (JEPA Policy), $48.38$\,M (Dual-independent) and $53.73$\,M (Dual-cross).
\item \textbf{SIGReg variant}: JEPA Policy plus an explicit isotropic-Gaussian regularizer on the encoder latents in the style of SIGReg~\cite{lejepa} (weight 0.09), used as an explicit-regularization reference in the collapse analysis.
\item \textbf{$\lambda_a=0$ variant}: JEPA Policy with the action loss weight set to zero, so the future objective trains the shared encoder alone; the anti-collapse control of Sec.~\ref{subsec:collapse}. It is trained on Square PH (seed 42) under an otherwise identical recipe and audited at its 220k-step EMA checkpoint.
\end{itemize}

\begin{table}[!htb]
\caption{Method properties of the compared approaches.}
\label{tab:main}
\centering
\begingroup
\scriptsize
\setlength{\tabcolsep}{2pt}
\resizebox{\columnwidth}{!}{%
\begin{tabular}{@{}lcccc@{}}
\toprule
Method & \shortstack{Few-step\\inference} & \shortstack{Future repr.\\supervision} & \shortstack{Shared\\attention} & \shortstack{Diagnostic\\signal} \\
\midrule
MIP (action-only) & Yes (two-step) & No & N/A & No \\
Diffusion Policy & No & No & N/A & No \\
Dual-independent$^{\dagger}$ & Yes (two-step) & Yes (separate) & No & Optional \\
Dual-cross & Yes (two-step) & Yes (cross-attn) & No & Optional \\
\textbf{JEPA Policy} & Yes (two-step) & Yes (shared) & Yes & Yes \\
\bottomrule
\end{tabular}
}
\vspace{1pt}
\parbox{\columnwidth}{\scriptsize $\dagger$ ACT-JEPA-style separate-branch topology; a controlled proxy rather than a faithful ACT-JEPA reproduction.}
\endgroup
\end{table}

\begin{figure}[!t]
\centering
\includegraphics[width=\columnwidth]{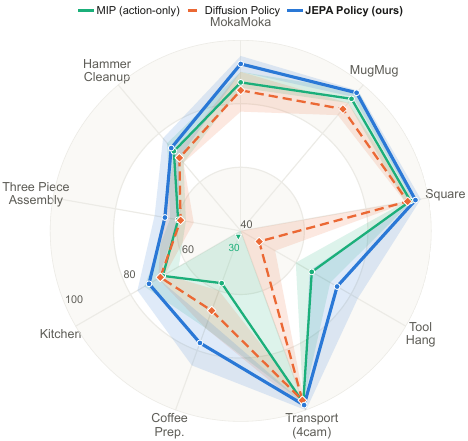}
\caption{
Main comparison over nine tasks (line: three-seed mean best-checkpoint success; band: seed range; radial axis $40$--$100$; means in Table~\ref{tab:main_results}).
A $\blacktriangledown$ and printed value mark a seed below the axis floor.
}
\label{fig:main_results}
\end{figure}

\begin{table*}[!t]
\caption{Best-checkpoint success rate (\%) on the nine tasks.}
\label{tab:main_results}
\centering
\footnotesize
\setlength{\tabcolsep}{2.6pt}
\rowcolors{2}{white}{gray!8}
\begin{tabular}{@{}l*{5}{rl}@{}}
\toprule
Task & \multicolumn{2}{c}{MIP (action-only)} & \multicolumn{2}{c}{Dual-indep.} & \multicolumn{2}{c}{Dual-cross} & \multicolumn{2}{c}{Diffusion Policy} & \multicolumn{2}{c}{\textbf{JEPA Policy}} \\
\midrule
MokaMoka & 86.7 & {\tiny 85/90/85} & 85.8 & {\tiny 82.5/87.5/87.5} & 82.5 & {\tiny 77.5/87.5/82.5} & 84.2 & {\tiny 85/77.5/90} & \textbf{92.5} & {\tiny 92.5/95/90} \\
MugMug & 94.2 & {\tiny 97.5/92.5/92.5} & 90.8 & {\tiny 87.5/90/95} & 86.7 & {\tiny 90/85/85} & 90.0 & {\tiny 90/87.5/92.5} & \textbf{96.7} & {\tiny 97.5/95/97.5} \\
Square & 94.2 & {\tiny 95/92.5/95} & 94.2 & {\tiny 95/92.5/95} & 94.2 & {\tiny 97.5/95/90} & 93.3 & {\tiny 92.5/95/92.5} & \textbf{95.8} & {\tiny 95/97.5/95} \\
Tool Hang & 65.8 & {\tiny 72.5/65/60} & 59.2 & {\tiny 75/52.5/50} & 68.3 & {\tiny 60/65/80} & 46.7 & {\tiny 47.5/52.5/40} & \textbf{75.0} & {\tiny 82.5/65/77.5} \\
Transport (4cam) & 97.5 & {\tiny 97.5/97.5/97.5} & \textbf{99.2} & {\tiny 97.5/100/100} & 97.5 & {\tiny 97.5/97.5/97.5} & 96.7 & {\tiny 95/95/100} & 98.3 & {\tiny 97.5/97.5/100} \\
Coffee Preparation & 57.5 & {\tiny 70/30/72.5} & 57.5 & {\tiny 62.5/47.5/62.5} & 61.7 & {\tiny 42.5/67.5/75} & 66.7 & {\tiny 72.5/60/67.5} & \textbf{77.5} & {\tiny 77.5/70/85} \\
Kitchen & 68.3 & {\tiny 67.5/62.5/75} & 72.5 & {\tiny 70/75/72.5} & 71.7 & {\tiny 72.5/67.5/75} & 69.2 & {\tiny 70/67.5/70} & \textbf{73.3} & {\tiny 72.5/77.5/70} \\
Three Piece Assembly & 60.0 & {\tiny 57.5/60/62.5} & 59.2 & {\tiny 65/52.5/60} & 63.3 & {\tiny 62.5/67.5/60} & 59.2 & {\tiny 65/57.5/55} & \textbf{64.2} & {\tiny 67.5/65/60} \\
Hammer Cleanup & 72.7 & {\tiny 76/68/74} & 68.7 & {\tiny 68/72/66} & 72.7 & {\tiny 74/74/70} & 70.0 & {\tiny 72/68/70} & \textbf{74.0} & {\tiny 74/70/78} \\
\midrule
Mean (9 tasks) & 77.4 & {\tiny 79.8/73.1/79.3} & 76.3 & {\tiny 78.1/74.4/76.5} & 77.6 & {\tiny 74.9/78.5/79.4} & 75.1 & {\tiny 76.6/73.4/75.3} & \textbf{83.0} & {\tiny 84.1/81.4/83.7} \\
\bottomrule
\end{tabular}

\vspace{2pt}
\parbox{\textwidth}{\footnotesize Large number: mean over seeds $\{41,42,43\}$. Small numbers: individual seeds in the same order. Matching seed indices use the same data ordering and initialization stream across arms. Bold values: best mean per task. The last row reports the nine-task mean and its per-seed values; means and gains are computed before rounding. The Transport topology cells use the aligned evaluation window from Sec.~\ref{subsec:exp_setup}.}

\end{table*}

\subsection{Main Results}
\label{subsec:main_results}

Table~\ref{tab:main_results} and Fig.~\ref{fig:main_results} present the main comparison on the nine LIBERO/robomimic/MimicGen tasks.
JEPA Policy improves over the MIP (action-only) arm on all nine tasks. The largest gains occur on Coffee Preparation ($57.5\rightarrow77.5$) and Tool Hang ($65.8\rightarrow75.0$), and the nine-task mean improves by $+5.6$ points ($77.4\rightarrow83.0$).
The Coffee Preparation baseline mean is depressed by an outlier seed (seed 42 reaches only 30.0 while seeds 41/43 reach 70.0/72.5); the JEPA Policy gain remains clear on every individual seed.
On Square, MugMug, Transport, and Hammer Cleanup, the gains are small relative to the evaluation granularity. On Hammer Cleanup, JEPA Policy reaches $74.0$, compared with $72.7$ for MIP, $70.0$ for Diffusion Policy, and $68.7/72.7$ for the dual controls.
The small numbers in Table~\ref{tab:main_results} provide the aggregate result for each seed. The weakest JEPA Policy seed ($81.4$ across tasks) remains above the strongest seed from every other arm ($79.8$ MIP, $78.1$ Dual-independent, $79.4$ Dual-cross, and $76.6$ Diffusion Policy).
This seed ordering shows that the nine-task mean is not produced by one favorable initialization of JEPA Policy or one unfavorable initialization of a comparison arm.
It does not turn three seeds into independent task replicates, which is why the statistical analysis below continues to use the task as its unit.

\begin{figure*}[!t]
\centering
\includegraphics[width=\textwidth]{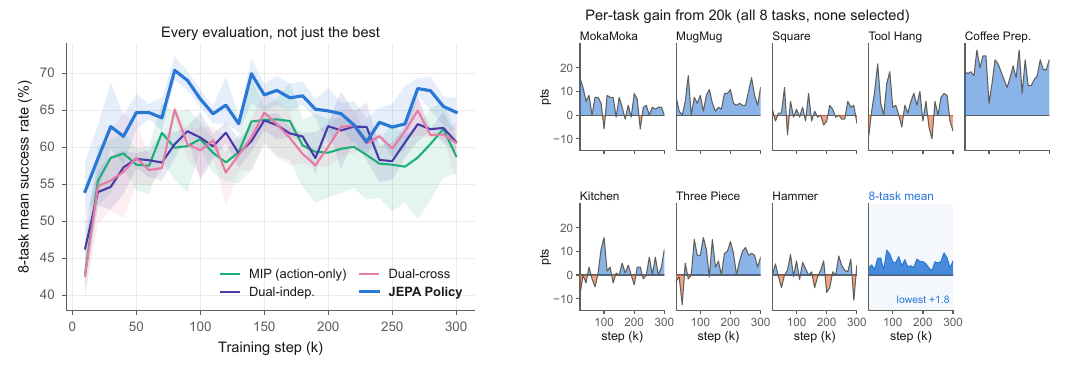}
\caption{
Training curves for eight tasks with three completed seeds in all four arms.
\emph{Left}: eight-task mean across 30 evaluations (line: seed mean; band: seed range).
\emph{Right}: per-task JEPA Policy gain over MIP from the second evaluation onward; blue/orange denotes positive/negative gain on a shared scale.
The eight-task mean is positive at all 29 evaluations (minimum $+1.8$); individual-task differences span $-12.5$ to $+27.5$ points.
}
\label{fig:learning}
\end{figure*}

\textbf{Statistical reading.}
We treat the task as the unit of analysis, since the three seeds of a task are not independent of one another.
Across the nine tasks, the mean gain over MIP is $+5.61$ points. The paired $t$ interval is $[+0.99,+10.23]$ ($t(8)=2.80$), and the paired bootstrap interval is $[+2.56,+9.82]$ ($20{,}000$ resamples). All nine task differences are positive, giving $p=0.0039$ for both the exact two-sided sign test and Wilcoxon signed-rank test.
Against Dual-cross, the mean is $+5.43$ points with an interval of $[+1.23,+9.63]$. All nine differences are positive (sign test and Wilcoxon $p=0.0039$).
Against Dual-independent, the mean is $+6.70$ points with an interval of $[+1.39,+12.02]$. Eight differences are positive and Transport is the exception (sign test $p=0.039$, Wilcoxon $p=0.012$).
We do not report a seed-level test. With three seeds, the paired statistic is $t(2)=4.18$ against a critical value of $4.303$. No three-seed design can reach $p<0.05$ under an exact two-sided sign-flip permutation, whose $p$-value floor is $0.25$.
The evidence for the effect is therefore its consistency across tasks, not its significance across seeds; Fig.~\ref{fig:learning} shows that it is also consistent across training time.

\textbf{The gain does not depend on the best-checkpoint protocol.}
Reporting the maximum over 30 evaluations is optimistic by construction, and the concern it raises is whether the reported gain is an artifact of that selection.
We therefore recomputed the same comparison from the stored evaluation curves under three protocols that involve no selection at all (Table~\ref{tab:protocol}).
The absolute success rates fall sharply under last-checkpoint reporting, with the nine-task MIP mean decreasing from $77.4$ to $62.5$. This confirms that best-checkpoint reporting is optimistic in \emph{level}.
The gain, however, remains between $+4.7$ and $+5.6$ points because both arms are selected by the identical rule and the selection bias largely cancels in the difference.
The last-checkpoint row uses one evaluation per task with no temporal averaging. Its per-task sign pattern drops to five of nine, whereas both averaging protocols retain eight of nine.
We retain best-checkpoint numbers in the main tables for comparison with the literature. Table~\ref{tab:protocol} shows that the conclusion does not depend on this reporting choice.

\begin{table}[!t]
\caption{\emph{Inference} cost of one policy decision (Tool Hang, seed 42, EMA weights, batch size 1) on a PPU-ZW810E accelerator.}
\label{tab:latency}
\centering
\footnotesize
\setlength{\tabcolsep}{3pt}
\begin{tabular}{@{}lrrrrr@{}}
\toprule
Method & \multicolumn{1}{c}{Params (M)} & \multicolumn{1}{c}{Passes} & \multicolumn{1}{c}{Mean (ms)} & \multicolumn{1}{c}{$p_{95}$ (ms)} & \multicolumn{1}{c}{Max Hz} \\
\midrule
MIP (action-only) & 43.2 & 2 & 12.9 & 13.4 & 77.3 \\
Diffusion Policy & 42.6 & 100 & 439.5 & 462.6 & 2.3 \\
\textbf{JEPA Policy} & 44.0 & 2 & 13.2 & 14.5 & 75.6 \\
\bottomrule
\end{tabular}

\vspace{2pt}
\parbox{\columnwidth}{\footnotesize Eager mode. JEPA Policy and MIP are timed interleaved in one PyTorch 2.6 process after 30 warmup and 300 synchronized iterations per arm; peak activation is 362/355\,MiB (JEPA/MIP). Instrumented entry points verify ``Passes.'' The full audit path, including the second future projection, costs $12.8$\,ms. Diffusion Policy uses the same task, device, and aligned-protocol checkpoint in its trained 100-step DDPM configuration, but runs in a separate PyTorch 2.9 process. This increases uncertainty in its absolute latency, not the $33\times$ gap. The parameter counts are capacity-matched.}

\end{table}

\begin{table}[!t]
\caption{Sensitivity of the nine-task mean success rate (\%) to the checkpoint reporting protocol.}
\label{tab:protocol}
\centering
\footnotesize
\setlength{\tabcolsep}{8pt}
\begin{tabular}{@{}lcccc@{}}
\toprule
Protocol & Baseline & \textbf{JEPA} & Gain & Improved \\
\midrule
Best of 30 evaluations & 77.4 & 83.0 & $+$5.6 & 9/9 \\
Mean of last 5 evaluations & 63.1 & 68.4 & $+$5.3 & 8/9 \\
Mean of all 30 evaluations & 62.8 & 67.5 & $+$4.7 & 8/9 \\
Last checkpoint only & 62.5 & 67.5 & $+$5.1 & 5/9 \\
\bottomrule
\end{tabular}

\vspace{2pt}
\parbox{\columnwidth}{\footnotesize Recomputed from the stored 30-evaluation curves of the same runs ($n=3$ seeds). ``Improved'' counts tasks with a positive three-seed mean gain. Values and gains are computed before rounding. The first row reproduces Table~\ref{tab:main_results}.}

\end{table}

\begin{figure}[!t]
\centering
\includegraphics[width=\columnwidth]{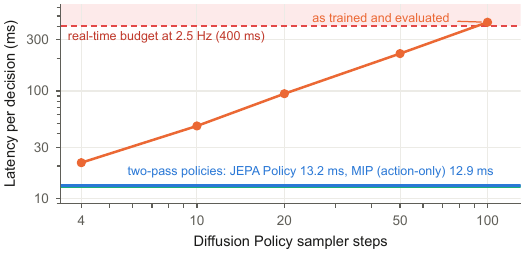}
\caption{
Decision latency against Diffusion Policy sampler length (Tool Hang, same accelerator, batch~1, log--log).
Shortened orange chains are timing-only; shading marks latency above the $400$\,ms budget.
JEPA Policy and MIP form one blue band ($13.2$ vs.\ $12.9$\,ms).
}
\label{fig:latency_steps}
\end{figure}

Against Diffusion Policy, JEPA Policy is ahead on all nine tasks, with a nine-task mean of $83.0$ versus $75.1$. Tool Hang accounts for much of this margin because Diffusion Policy reaches only $46.7$ on that task. Excluding Tool Hang gives a more conservative mean difference of $+5.4$ points. The methods are closest on Transport ($98.3$ vs.\ $96.7$), Square ($95.8$ vs.\ $93.3$), and Hammer Cleanup ($74.0$ vs.\ $70.0$).
The comparison therefore supports two distinct readings.
Success is comparable or better under the evaluated task and training configurations, but the magnitude of the average gain is task-dependent.
The latency comparison is more uniform because it follows directly from two policy passes rather than a 100-step denoising chain.
\textbf{Inference cost.}
Table~\ref{tab:latency} measures hardware latency and verifies the two forward passes; the second control pass skips the future projection.
JEPA Policy averages $13.2$\,ms ($p_{95}=14.5$\,ms), using $3\%$ of the $400$\,ms budget and sustaining $76$\,Hz.
Relative to MIP, the future branch adds $0.79$\,M parameters, $0.29$\,ms ($+2.2\%$), and $7$\,MiB of peak activation memory.
The trained 100-step Diffusion Policy uses 100 verified calls and takes $439.5$\,ms ($2.3$\,Hz), or $33\times$ longer, at a similar parameter count.
Shortening it to $50/20/10/4$ steps gives $223.0/94.4/47.3/21.5$\,ms (Fig.~\ref{fig:latency_steps}); these are timing-only settings, and four steps remain $1.6\times$ slower.
Per-call costs are comparable ($\approx4.4$\,ms per denoising step and $\approx6.6$\,ms per JEPA Policy pass); the difference is the call count.
These single-task, single-seed timings establish only the latency scale.

\looseness=-1 \textbf{Training and evaluation cost.}
Training is where the future branch incurs cost.
For the same $300{,}000$ steps, it raises optimization time by $12$--$22\%$ ($0.222\rightarrow0.271$\,s/step on Tool Hang).
The 30 evaluations add $2.7$--$5.2$\,h, up to one fifth of a run (Table~\ref{tab:traincost}).
A JEPA Policy step costs about $1.8\times$ a Diffusion Policy step (median over the original eight-task timing audit; $1.5$--$2.1\times$ on the three shown) because it performs two passes and encodes the future.
This higher optimization cost accompanies a $33\times$ cheaper decision.
At one seed, JEPA Policy reaches a shared success target in fewer steps than each comparison arm on six of the eight tasks in this timing-to-target audit, but the direction is not significant ($p=0.29$).
Every run uses one accelerator, takes about one day, and requires no video pretraining.
The dual variants were about twice as slow in a different launch batch, so that observation is uncontrolled.

\begin{table}[!htb]
\caption{Training and evaluation cost of one run ($300{,}000$ gradient steps, seed 42) on the accelerator of Table~\ref{tab:latency}.}
\label{tab:traincost}
\centering
\footnotesize
\setlength{\tabcolsep}{4pt}
\begin{tabular}{@{}llcccc@{}}
\toprule
Task & Arm & s/step & Train (h) & Eval (h) & Total (h) \\
\midrule
\multirow{3}{*}{Tool Hang} & Diffusion Policy & 0.136 & 11.3 & N/A & N/A \\
 & MIP (action-only) & 0.222 & 18.5 & 4.8 & 23.3 \\
 & \textbf{JEPA Policy} & 0.271 & 22.6 & 4.8 & 27.4 \\
\midrule
\multirow{3}{*}{Square} & Diffusion Policy & 0.136 & 11.4 & N/A & N/A \\
 & MIP (action-only) & 0.236 & 19.7 & 2.7 & 22.4 \\
 & \textbf{JEPA Policy} & 0.289 & 24.0 & 2.7 & 26.7 \\
\midrule
\multirow{3}{*}{MokaMoka} & Diffusion Policy & 0.206 & 17.2 & N/A & N/A \\
 & MIP (action-only) & 0.273 & 22.7 & 5.2 & 27.9 \\
 & \textbf{JEPA Policy} & 0.305 & 25.5 & 5.1 & 30.6 \\
\bottomrule
\end{tabular}

\vspace{2pt}
\parbox{\columnwidth}{\footnotesize ``Train'' is run wall-clock time minus the 30 periodic rollout evaluations and therefore measures optimization only. ``Eval'' is the subtracted evaluation time, and ``Total'' is their sum. ``s/step'' is training time divided by $300{,}000$. The Diffusion Policy logger records per-step optimization time but not evaluation duration, so its Eval and Total entries are marked N/A. Because it uses a different codebase, its s/step is indicative rather than a controlled measurement.}

\end{table}

\begin{figure*}[t]
\centering
\includegraphics[width=\textwidth]{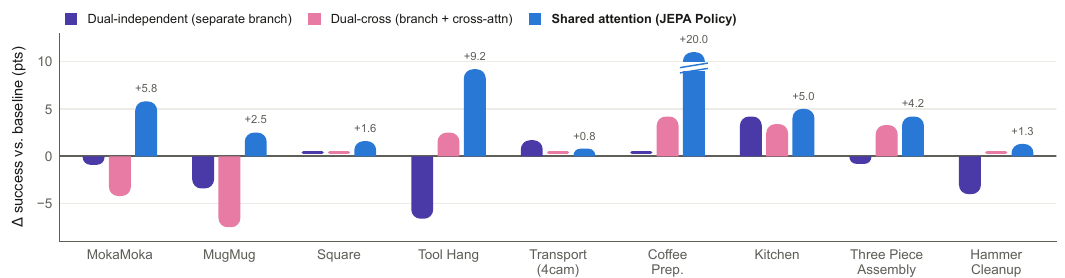}
\caption{
Gain over MIP (three-seed mean best-checkpoint difference; per-task means in Table~\ref{tab:main_results}).
Dashes mark ties; the broken Coffee Preparation bar is $+20.0$; Transport topology values use the aligned window defined in Sec.~\ref{subsec:exp_setup}.
JEPA Policy averages $+5.6$ points across nine improvements, versus $-1.1$ and $+0.2$ for the dual variants.
}
\label{fig:shared_dual}
\end{figure*}

\subsection{Shared Attention vs.\ Separate Branches}
\label{subsec:future_ablation}

We test whether future prediction must share the action-generating stack to improve the policy.
Figure~\ref{fig:shared_dual} compares JEPA Policy with two dual-branch implementations that retain the training recipe, losses, and parameter protocol while changing only the topology.
Dual-independent places the separate-branch topology of ACT-JEPA~\cite{act_jepa} within our training recipe. It retains the same losses and MIP training schedule, while using stopped targets instead of an EMA target encoder.
It is therefore a controlled proxy for the separate-branch \emph{design choice}, not a faithful reproduction of ACT-JEPA. We remove differences in the original objective and EMA machinery so that the comparison isolates topology.

Routing the future loss through a separate branch does not reproduce the gain.
Averaged over the nine tasks, Dual-independent reaches $76.3$ and Dual-cross reaches $77.6$, compared with $77.4$ for MIP and $83.0$ for JEPA Policy. The two dual variants remain close to MIP, while the shared stack is $5.4$--$6.7$ points higher than all three.
JEPA Policy is the sole best variant on eight tasks. The exception is Transport (4cam), where Dual-independent reaches $99.2$ and JEPA Policy reaches $98.3$. This difference is smaller than one evaluation episode ($2.5$ points) in the aligned 10k--230k window.
At the task level, Dual-independent is at or below MIP on seven of nine tasks and Dual-cross on five of nine. The separate-branch variants therefore match the baseline overall.
Adding the future loss outside the shared stack yields no mean improvement, and cross-attention between the branches does not recover the gain.
Two properties help interpret this result.
First, both have \emph{more} trainable parameters than JEPA Policy ($48.38$\,M and $53.73$\,M against $43.96$\,M, Sec.~\ref{subsec:baselines}) and neither beats the MIP (action-only) arm, so the gain cannot be attributed to added capacity.
Second, Dual-cross gives the two token groups mutual visibility through cross-attention at every layer but still provides no gain. Mutual visibility alone is therefore insufficient.
JEPA Policy also uses the same parameters to produce both readouts. Section~\ref{subsec:mechanism} measures how this difference changes the gradient reaching the action pathway.
These results identify joint processing of action and future tokens in one Transformer as the important architectural component. A future-prediction branch outside the shared stack does not provide the same effect.
This conclusion is stronger than observing that JEPA Policy has an additional loss.
The dual controls retain that loss, use more parameters, and expose both branches to the same observation memory, yet their mean performance remains near MIP.
The controlled difference is where the future objective enters the action-generating computation.
The aligned Transport window preserves every reported maximum in this comparison; excluding Transport entirely also leaves the ordering unchanged.

\subsection{Where the Future Objective Acts}
\label{subsec:mechanism}

\begin{figure}[!htb]
\centering
\includegraphics[width=\columnwidth]{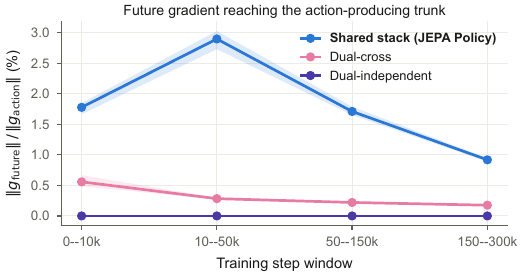}
\caption{
\looseness=-1 Norm of the future-loss gradient reaching the trunk that produces the action, relative to the action-loss gradient on the same parameters, averaged over training windows (line: mean over seeds $\{41,42,43\}$; band: seed range; Tool Hang).
In Dual-independent the quantity is exactly zero at every logged step, by construction.
The shared stack stays an order of magnitude above Dual-cross throughout, peaking in mid-training rather than decaying.
}
\label{fig:mechanism}
\end{figure}

The ablation above shows \emph{that} the topology matters; this subsection reports what the three topologies do to the gradient, which is the only channel through which the future objective can change the action parameters.
We record, at every logged training step, the norm of the future-loss gradient on the parameters that produce the action readout, relative to the action-loss gradient on those same parameters.

Three findings distinguish the arms (Fig.~\ref{fig:mechanism}).
In Dual-independent, the ratio is zero at every step and seed because the decoder stacks share no parameters. The future loss still reaches the shared conditioning path, so this result concerns the action decoder rather than the entire representation. This property makes Dual-independent a direct control for topology.
In Dual-cross the ratio is non-zero but small, and it decays monotonically from $0.5\%$ to $0.2\%$; cross-attention does connect the branches, but the connection thins out as training proceeds.
In the shared stack, the ratio rises to $2.9\%$ in the $10$--$50$k window and remains $0.9\%$ at $300$k. It is five to ten times the Dual-cross ratio in every window, with little variation among seeds.
At convergence the same ordering holds on a fixed evaluation manifest: $0.68\%$ for the shared stack against $0.135\%$ for Dual-cross and exactly $0$ for Dual-independent.

Two observations limit this interpretation.
First, mutual visibility alone does not explain the ranking. The action-loss gradient on the Dual-cross future tokens is about five times larger than in the shared stack, yet Dual-cross shows no gain. The relevant direction is therefore the future objective reaching the action parameters, not the reverse.
Second, we do not report a gradient \emph{alignment} statistic. The cosine between the two gradients is small ($|\cos|<0.03$) and its sign is not stable across measurement conditions, so we treat only the magnitude as evidence.
The measurement is a share of gradient norm, not of the optimizer's update. AdamW can amplify or attenuate a persistent component, so these values should not be read as update shares.

\begin{table}[!t]
\caption{Cutting each route between the future objective and the action, at a single seed.}
\label{tab:paths}
\centering
\footnotesize
\setlength{\tabcolsep}{5pt}
\begin{tabular}{@{}lrrrr@{}}
\toprule
Task & \multicolumn{1}{c}{MIP} & \multicolumn{1}{c}{Back-cut} & \multicolumn{1}{c}{Fwd-cut} & \multicolumn{1}{c}{\textbf{JEPA Policy}} \\
\midrule
MokaMoka & 90.0 & 100.0 & 90.0 & 95.0 \\
MugMug & 92.5 & 95.0 & 92.5 & 95.0 \\
Square & 92.5 & 97.5 & 97.5 & 97.5 \\
Tool Hang & 65.0 & 17.5 & 55.0 & 65.0 \\
Transport (4cam) & 97.5 & 95.0 & 97.5 & 97.5 \\
Coffee Preparation & 30.0 & 65.0 & 70.0 & 70.0 \\
Kitchen & 62.5 & 62.5 & 72.5 & 77.5 \\
Three Piece Assembly & 60.0 & 52.5 & 52.5 & 65.0 \\
\midrule
Mean (8 tasks) & 73.8 & 73.1 & 78.4 & \textbf{82.8} \\
\bottomrule
\end{tabular}

\vspace{2pt}
\parbox{\columnwidth}{\footnotesize All four arms use seed 42 and the same code base, so the comparison is seed-matched. These numbers are \emph{not} comparable with the three-seed means in Table~\ref{tab:main_results}. \emph{Back-cut} preserves attention from action tokens to the future token but detaches the future loss from the shared trunk. \emph{Fwd-cut} allows the future loss to shape the shared representation but removes the future token from the action tokens' attention. The Transport cells use the aligned 10k--230k window; all maxima occur before its boundary.}

\end{table}

\textbf{Cutting each route.} The gradient shares above are correlational: they show that the future objective reaches the action parameters, not that the gain depends on it. We therefore severed each route directly and ran a matched training and evaluation comparison at one seed (Table~\ref{tab:paths}).
Severing the backward route removes the aggregate gain. Back-cut averages $73.1$, compared with $73.8$ for MIP at the same seed and $82.8$ for the full method. A future head whose gradient cannot reach the shared trunk therefore provides no aggregate improvement. Severing the forward route retains about half of the gain ($78.4$).
Together with Fig.~\ref{fig:mechanism}, this experiment shows where the future objective acts. Most of the gain depends on its gradient reaching the shared representation. Direct attention from action tokens to the future token provides the remaining improvement.

\begin{table}[t]
\caption{Future-loss ratio (fixed $k=4$, i.e., $0.2$\,s at $20$\,Hz) and horizon (fixed $\rho=0.10$) ablations (best-checkpoint success rate \%, three-seed mean $\pm$ std).}
\label{tab:ratio_horizon}
\centering
\scriptsize
\setlength{\tabcolsep}{3pt}
\resizebox{\columnwidth}{!}{%
\begin{tabular}{@{}l*{5}{r@{}l}@{}}
\toprule
Config. & \multicolumn{2}{c}{MokaMoka} & \multicolumn{2}{c}{MugMug} & \multicolumn{2}{c}{Square} & \multicolumn{2}{c}{Tool Hang} & \multicolumn{2}{c}{Transport (4cam)} \\
\midrule
baseline & 86.7 & $\,\pm\,$2.9 & 94.2 & $\,\pm\,$2.9 & 94.2 & $\,\pm\,$1.4 & 65.8 & $\,\pm\,$6.3 & 97.5 & $\,\pm\,$0.0 \\
$\rho=0.05$ & \textbf{93.3} & $\,\pm\,$5.2 & 92.5 & $\,\pm\,$2.5 & 95.0 & $\,\pm\,$0.0 & 70.8 & $\,\pm\,$8.0 & 99.2 & $\,\pm\,$1.4 \\
$\rho=0.10$ & 92.5 & $\,\pm\,$2.5 & 96.7 & $\,\pm\,$1.4 & 95.8 & $\,\pm\,$1.4 & \textbf{75.0} & $\,\pm\,$9.0 & 98.3 & $\,\pm\,$1.4 \\
$\rho=0.20$ & 91.7 & $\,\pm\,$1.4 & \textbf{97.5} & $\,\pm\,$2.5 & 97.5 & $\,\pm\,$0.0 & 70.8 & $\,\pm\,$8.0 & 99.2 & $\,\pm\,$1.4 \\
$k=2$ & 90.0 & $\,\pm\,$0.0 & 94.2 & $\,\pm\,$2.9 & 95.0 & $\,\pm\,$0.0 & 69.2 & $\,\pm\,$13.8 & 99.2 & $\,\pm\,$1.4 \\
$k=6$ & \textbf{93.3} & $\,\pm\,$2.9 & 95.0 & $\,\pm\,$2.5 & \textbf{98.3} & $\,\pm\,$2.9 & 70.8 & $\,\pm\,$6.3 & \textbf{100.0} & $\,\pm\,$0.0 \\
\bottomrule
\end{tabular}
}
\end{table}

This probe has two limitations. It uses one seed, so it supports only the aggregate ordering rather than a stable effect size. Each 40-episode evaluation also has a $2.5$-point resolution, and Table~\ref{tab:main_results} shows substantial per-task variation across seeds.
Back-cut matches the aggregate baseline through large offsetting changes: $-47.5$ points on Tool Hang and $+35.0$ on Coffee Preparation. The seed-42 Coffee baseline is the collapsed run.
We therefore claim only that severing the backward route removes the aggregate gain and destabilizes per-task behavior. We attach no significance statement to a single seed.

\subsection{Future-Loss Ratio and Horizon}
\label{subsec:ratio_horizon}
\begin{figure*}[t]
\centering
\includegraphics[width=\textwidth]{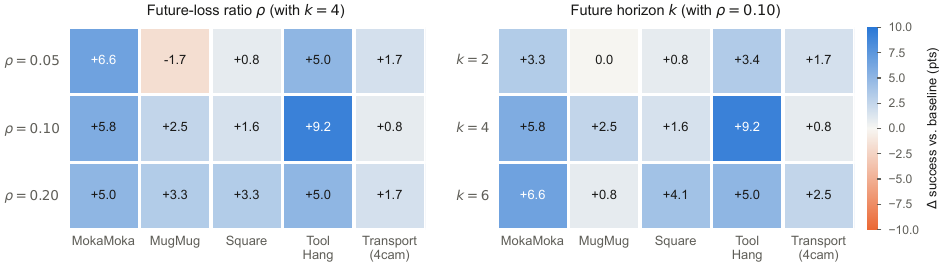}
\caption{
Hyperparameter sensitivity as gain over the baseline (three-seed means, in points): future-loss ratio $\rho$ (left) and future horizon $k$ (right); blue is better than the baseline, orange worse.
No configuration is materially worse than the baseline; $\rho\in[0.1,0.2]$ and $k\in\{4,6\}$ are uniformly strong choices.
}
\label{fig:ablation}
\end{figure*}

Table~\ref{tab:ratio_horizon} and Fig.~\ref{fig:ablation} sweep the two hyperparameters of the future branch.
No configuration with future supervision is materially worse than the baseline: across the five tasks and six settings, the only cell below the baseline is $\rho=0.05$ on MugMug ($92.5$ vs.\ $94.2$), a $1.7$-point difference smaller than one evaluation episode. The method is otherwise insensitive within the swept range.
$\rho=0.10$--$0.20$ dominates $\rho=0.05$ on the precision-heavy tasks (MugMug, Square, Tool Hang), and $k=4$--$6$ dominates $k=2$, consistent with the selection principle of Sec.~\ref{subsec:problem}: too short a horizon makes the future target nearly redundant with the current observation.
Tool Hang is the most sensitive task (best at $\rho=0.10$, $k=4$), while Square and Transport favor the larger $k=6$.

\subsection{Robustness to the Observation Configuration}
\label{subsec:cameras}

Table~\ref{tab:cameras} varies the camera configuration at seed 42.
With wrist images only, six of seven tasks improve and one ties; MokaMoka and Coffee Preparation each gain $10.0$ points.
With fixed cameras only, all five usable differences remain within $\pm5$ points, so the gain is not reproduced.
These results are directional: an identical single-seed configuration varied by 17.5 points between launch batches.
The fixed-camera results therefore provide insufficient evidence of a reversal. The main comparison instead averages three seeds (Sec.~\ref{subsec:main_results}).

\begin{table}[t]
\caption{Camera-configuration sweep (best-checkpoint success rate \%, single seed 42; directional only).}
\label{tab:cameras}
\centering
\scriptsize
\setlength{\tabcolsep}{2.4pt}
\resizebox{\columnwidth}{!}{%
\begin{tabular}{@{}lrrrrrrr@{}}
\toprule
Config. & \multicolumn{1}{c}{Moka} & \multicolumn{1}{c}{Mug} & \multicolumn{1}{c}{Square} & \multicolumn{1}{c}{\shortstack{Tool\\Hang}} & \multicolumn{1}{c}{Coffee} & \multicolumn{1}{c}{Kitchen} & \multicolumn{1}{c}{\shortstack{Three\\Piece}} \\
\midrule
Standard, baseline & 90.0 & 92.5 & 92.5 & 65.0 & 30.0 & 62.5 & 60.0 \\
Standard, future-4 & \textbf{95.0} & \textbf{95.0} & \textbf{97.5} & \textbf{65.0} & \textbf{70.0} & \textbf{77.5} & \textbf{65.0} \\
Wrist only, baseline & 62.5 & 80.0 & 80.0 & 27.5 & 37.5 & 42.5 & 15.0 \\
Wrist only, future-4 & 72.5 & 87.5 & 80.0 & 30.0 & 47.5 & 47.5 & 17.5 \\
Fixed only, baseline & 87.5 & 70.0 & 90.0 & 35.0 & 67.5 & 65.0 & 42.5 \\
Fixed only, future-4 & N/A & N/A & 87.5 & 37.5 & 62.5 & 62.5 & 40.0 \\
\bottomrule
\end{tabular}}

\vspace{2pt}
\parbox{\columnwidth}{\footnotesize Transport is omitted because the camera sources differ across rows. The fixed-only future-4 LIBERO cells are marked N/A because a data bug invalidated that batch.}

\end{table}

\subsection{Alternative Designs and Negative Results}
\label{subsec:negative}

\textbf{Explicit isotropic regularization is fragile.}
Adding the SIGReg isotropic-Gaussian objective~\cite{lejepa} (weight 0.09) lowers performance on all five tasks (Table~\ref{tab:sigreg}), with MokaMoka at 0.0 across all seeds despite normal training completion.
The prior may over-regularize features already constrained by action supervision; without a weight sweep, this mechanism remains a hypothesis and does not imply that SIGReg generally fails.


\textbf{State inputs.}
With low-dimensional state inputs instead of images (three seeds), the effect is task-dependent: Transport improves markedly ($52.5\rightarrow62.5$) while Tool Hang dips within one granularity step ($75.0\rightarrow72.5$).
Future supervision thus systematically harms neither observation modality, and its benefit is not exclusive to image inputs.

\begin{table}[!b]
\caption{Adding an explicit SIGReg regularizer (weight 0.09) to JEPA Policy (best-checkpoint success rate \%, three-seed mean $\pm$ std).}
\label{tab:sigreg}
\centering
\scriptsize
\setlength{\tabcolsep}{3pt}
\resizebox{\columnwidth}{!}{%
\begin{tabular}{@{}l*{5}{r@{}l}@{}}
\toprule
Config. & \multicolumn{2}{c}{MokaMoka} & \multicolumn{2}{c}{Square} & \multicolumn{2}{c}{Tool Hang} & \multicolumn{2}{c}{Coffee} & \multicolumn{2}{c}{Three Piece} \\
\midrule
\textbf{JEPA Policy} & \textbf{92.5} & $\,\pm\,$2.5 & \textbf{95.8} & $\,\pm\,$1.4 & \textbf{75.0} & $\,\pm\,$9.0 & \textbf{77.5} & $\,\pm\,$7.5 & \textbf{64.2} & $\,\pm\,$3.8 \\
$+$ SIGReg ($w{=}0.09$) & 0.0 & $\,\pm\,$0.0 & 42.5 & $\,\pm\,$9.0 & 2.5 & $\,\pm\,$0.0 & 12.5 & $\,\pm\,$11.5 & 36.7 & $\,\pm\,$2.9 \\
\bottomrule
\end{tabular}
}
\end{table}

\subsection{Representation Collapse Audit}
\label{subsec:collapse}

We evaluated collapse with a pre-specified spectral protocol. A model can avoid a constant output yet still lose rank as part of its embedding spectrum approaches zero~\cite{dim_collapse}. A test that only checks whether outputs differ would miss this failure mode.
The audit covers JEPA Policy and matched action-only controls on seven tasks and three seeds, for 42 runs in total. We evaluate the best and latest checkpoint of each run, giving 84 checkpoints and 168 representation matrices over fixed 4{,}096-sample manifests.
All metrics are computed in float64. They include effective rank, top-eigenvalue share, and tail energy TE32 with a pre-specified low-rank boundary of $0.05$. A fixed 256-candidate retrieval test measures whether predictor outputs depend on the input.
TE32 is the share of centered spectral energy outside the leading 32 directions. Let $X_c\in\mathbb{R}^{N\times D}$ be the mean-centered representation matrix and $\lambda_1\ge\dots\ge\lambda_D$ the eigenvalues of $X_c^\top X_c$. Then $\mathrm{TE32}=1-\sum_{i\le 32}\lambda_i/\sum_i\lambda_i$. Small values indicate concentration in few directions. The pre-specified boundary $\mathrm{TE32}=0.05$ marks the point at which 32 directions contain $95\%$ of the centered energy.

\begin{figure}[!htb]
\centering
\includegraphics[width=\columnwidth]{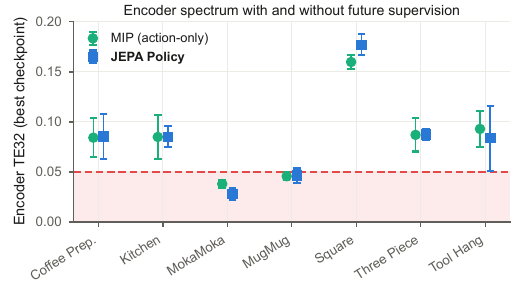}
\caption{
Encoder TE32 at the best checkpoint (marker: three-seed mean; bar: worst seed; shading: below the pre-specified $0.05$ boundary).
JEPA Policy tracks MIP, including the MokaMoka and MugMug boundary exceptions.
}
\label{fig:spectra}
\end{figure}

The outcome supports a bounded claim: there is \emph{no evidence of complete collapse or of systematic contraction to the pre-specified low-rank boundary}, and predictor retrieval is far above chance (input-dependent predictions); Fig.~\ref{fig:spectra} shows representative spectra.
The pre-specified task-level exception rule did trigger. Encoder TE32 for MugMug and MokaMoka, and predictor TE32 for Tool Hang and Square, crossed or touched the boundary at specific endpoints. We therefore do not claim that every task avoids low-rank contraction.
A non-collapsed point estimate alone would not establish that the action loss supplies the constraint, because the same optimization might remain stable without it.
For this reason, the audit and the $\lambda_a=0$ control answer complementary questions.
The multi-task audit tests whether collapse occurs under the complete training objective, while the removal control tests what changes when action supervision is absent.
A complementary trajectory experiment used four 50k-step runs with snapshots at $\{0,1\mathrm{k},5\mathrm{k},10\mathrm{k},25\mathrm{k},50\mathrm{k}\}$ steps. During the first 1{,}000 steps, the fixed future loss fell by 97.7\% and all seeds crossed the low-rank boundary. TE32 then partially recovered instead of decreasing monotonically. Retrieval Recall@5 remained far above chance at every endpoint (permutation $p=1/1001$).

\textbf{Removing the action loss collapses the representation.} The audit above shows only that collapse does not occur when the action loss is present.
We scored the $\lambda_a=0$ variant from Sec.~\ref{subsec:baselines} under the same protocol. This variant receives no action-loss gradient, while the future loss still trains the encoder and shared trunk.
Encoder TE32 falls from $0.170$ to $0.0014$, about $36\times$ below the pre-specified $0.05$ boundary. Centered effective rank decreases from $181$ to $24$ of $384$. In addition, $92\%$ of the embedding energy moves into the shared mean. The cosine similarity between two samples rises to $0.918$, compared with $0.021$ in the matched normal run.
The collapse is not complete because predictor retrieval remains far above chance (Recall@1 $0.976$, permutation $p=0.001$). This value is higher than the $0.525$ of the normal run, consistent with a future representation that becomes nearly deterministic within a 24-dimensional subspace.
The future loss alone therefore admits a degenerate solution in this experiment, whereas the action loss prevents it. This control covers only one task and one seed.

\begin{table}[!t]
\caption{Rollout audit: early Future-4 NMSE vs.\ failure (24 checkpoints; 5{,}200 episodes). Episode-resampled confidence intervals are descriptive conditional on these checkpoints.}
\label{tab:auroc}
\centering
\scriptsize
\setlength{\tabcolsep}{3.2pt}
\resizebox{\columnwidth}{!}{%
\begin{tabular}{@{}lcccc@{}}
\toprule
Task & Episodes & Failure rate & AUROC [95\% CI] & PR-AUC \\
\midrule
MugMug & 600 & 14.5\% & \textbf{0.754} [0.693, 0.810] & 0.377 \\
Tool Hang & 600 & 30.8\% & 0.722 [0.676, 0.766] & 0.532 \\
Coffee Preparation & 600 & 38.8\% & 0.662 [0.617, 0.706] & 0.566 \\
Three Piece Assembly & 600 & 47.7\% & 0.624 [0.578, 0.667] & 0.588 \\
Transport & 1{,}000 & 7.3\% & 0.623 [0.566, 0.681] & 0.127 \\
Square & 600 & 10.8\% & 0.620 [0.544, 0.694] & 0.167 \\
Kitchen & 600 & 21.8\% & 0.572 [0.513, 0.629] & 0.322 \\
MokaMoka & 600 & 16.7\% & 0.554 [0.488, 0.618] & 0.254 \\
\bottomrule
\end{tabular}
}
\end{table}

\begin{figure}[!t]
\centering
\includegraphics[width=\columnwidth]{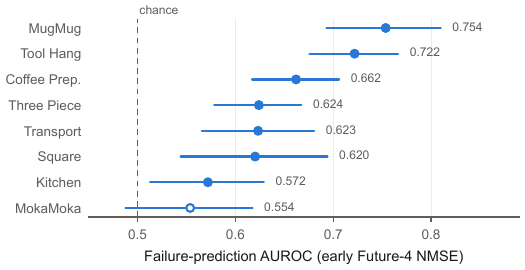}
\caption{
Per-task failure-prediction AUROC of the early Future-4 NMSE with conditional episode-level 95\% confidence intervals.
The open marker (MokaMoka) denotes a confidence interval that crosses chance.
}
\label{fig:auroc}
\end{figure}

\subsection{Future-Consistency Diagnostic at Rollout}
\label{subsec:diag_results}

We evaluated the diagnostic from Sec.~\ref{subsec:future_consistency} on 24 frozen checkpoints from eight tasks and three seeds. Checkpoint selection used no audit metric. The evaluation contains 5{,}200 rollout episodes, including 4{,}040 successes and 1{,}160 failures.
Before analysis, we verified byte-identical episode evidence across worker counts after correcting a reset-seed bug.
Each episode is scored by the median Future-4 NMSE (a $0.2$\,s-ahead prediction; Sec.~\ref{subsec:exp_setup}) of its first five policy decisions, i.e., by whether the future prediction is already inconsistent \emph{early} in the episode.
For one decision, the NMSE is $d_{\mathrm{NMSE}}(\hat{\mathbf{z}}^{(1)}_{t+k},\mathbf{z}_{t+k})$ as defined in Sec.~\ref{subsec:joint_loss}.
The normalization makes episodes comparable within a task despite different latent scales. An episode enters the analysis only when its first five decisions all have valid future targets.
The $95\%$ intervals use $10{,}000$ episode-level percentile-bootstrap resamples with a fixed seed. They are descriptive conditional on the three evaluated checkpoints per task, not checkpoint-population inference. Episode-level two-sided permutation tests use $10{,}000$ resamples; checkpoint robustness is reported separately below.

\textbf{The error ranks failures.}
Pooled within each task, failures have higher early NMSE than successes on all eight tasks. AUROC exceeds $0.5$ throughout, and seven conditional episode-level intervals exclude chance (Table~\ref{tab:auroc}, Fig.~\ref{fig:auroc}). MugMug (0.754) and Tool Hang (0.722) have the highest AUROC values. Only the MokaMoka interval crosses chance. On MugMug, failures outrank successes in $75.4\%$ of random failure--success pairs.
\mbox{At the checkpoint level,} 22 of 24 point estimates exceed 0.5 and 23 of 24 leave-one-seed-out tests agree in direction.
The pooled Transport value is not stable at the checkpoint level because two of its three seeds are individually below 0.5.

We use threshold-independent AUROC, following established protocols for misclassification and out-of-distribution detection~\cite{msp_ood}. The computational cost differs from ensemble uncertainty. Our score comes from a policy forward pass that is already required, whereas an ensemble needs $M$ trained networks and $M$ passes per decision~\cite{deep_ensembles}.
AUROC measures whether a randomly selected failure tends to receive a larger score than a randomly selected success.
It does not choose an alarm threshold and does not imply that the same numerical score has the same meaning across tasks.
This distinction is central here because the predictor is useful for within-task ranking even when absolute error scales differ.

\textbf{Two controls rule out trivial explanations.}
Residualizing the NMSE against the magnitude of realized latent visual change leaves the signal unchanged (checkpoint-macro AUROC $0.6184\rightarrow0.6183$). The association is therefore not explained by failures producing more motion.
A persistence baseline that sets the future latent equal to the current latent is near chance (0.508 vs.\ 0.618 for the learned predictor). The signal therefore depends on the learned prediction rather than generic latent motion alone.

\textbf{Raw scores must be conditioned per task.}
Pooling raw NMSE across tasks removes the signal (AUROC 0.501, CI [0.484, 0.519]; permutation $p=0.68$). This is a Simpson-type effect caused by task-specific error scales.
Any monitoring application must therefore calibrate per task (and for some tasks per checkpoint).
Post-hoc Youden thresholds are descriptive because they are chosen and evaluated on the same data. MugMug transfers best across seeds (held-out balanced accuracy 0.69--0.76), and Coffee transfers moderately. Kitchen and MokaMoka support no useful raw threshold. Tool Hang and Transport show usable ranking but a drifting absolute scale.
We use the diagnostic as a per-task-calibrated risk signal for failure analysis and human hand-off, not as a failure probability. It is learned from successful demonstrations and requires no foundation model, separate scoring network, or extra action samples. Dedicated runtime monitors use a different design~\cite{sentinel,fail_detect}. Operational thresholds still require held-out rollouts labeled by outcome. Because we ran no direct comparison, our claims concern cost and signal type rather than detection accuracy.
Thus, the positive AUROC result establishes an ordering signal, whereas deployment requires a second calibration stage on outcome-labeled data from the intended task and hardware.
Keeping these two stages separate avoids turning a retrospective diagnostic into an unsupported claim of calibrated online failure prediction.

\subsection{Real-Robot Evaluation}
\label{subsec:real_robot}
\begin{figure*}[!t]
\centering
\includegraphics[width=\textwidth]{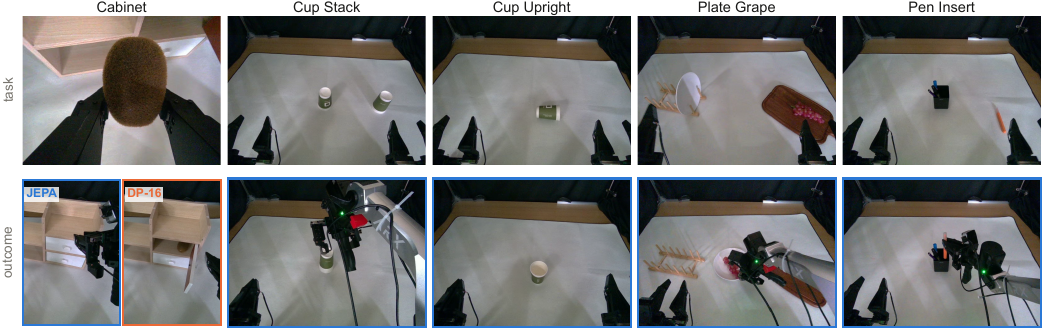}
\caption{
Representative task and outcome frames from successful JEPA Policy episodes on the five real-robot tasks. The Cabinet task frame uses the left-wrist camera to show the kiwi grasp; the remaining frames use the base camera. The Cabinet outcome cell additionally contrasts the JEPA Policy success (door closed) with a DP-16 terminal failure (door left open). Policies receive all views resized to $128\times128$.
}
\label{fig:real_tasks}
\end{figure*}
\looseness=-1
We repeated the comparison on a physical two-arm ARX-5; one NVIDIA GeForce RTX 3090 ran inference from three $128\times128$ RGB streams, with the driver at $30$\,Hz and actions at $10$\,Hz.
Across $3{,}048$ steady-state JEPA Policy chunks from $160$ episodes (first chunk excluded), frame-to-chunk latency was $20.1/63.3$\,ms (median/$p_{95}$).
MIP/DP-16/DP-100 recorded $27.6/67.5$, $79.2/130.8$ and $316.2/373.3$\,ms (median/$p_{95}$).
All methods used a synchronous rollout loop: each policy call completed before its returned action chunk was executed, and action execution did not overlap the next policy call.
JEPA Policy's $20.1$\,ms median latency was well below the $100$\,ms action period, so it required no asynchronous action queue.
JEPA Policy and MIP each executed eight actions per chunk, whereas DP-16/DP-100 skipped one/three expired positions, leaving eight/six executable actions.
The five tasks are Cabinet, Cup Stack, Cup Upright, Plate Grape, and Pen Insert (Fig.~\ref{fig:real_tasks}); their arm assignments are reported with Table~\ref{tab:real_robot}.
For each task, all compared methods were trained from scratch on the same 100 expert demonstrations.
Success required completing the operator-confirmed task goal within a fixed task-specific action-chunk budget shared by all methods; wall-clock duration, inference latency, and control-loop timing did not affect the label.
Each policy was evaluated at $60$k, $100$k, and $140$k steps for $10$ episodes per checkpoint.
Diffusion Policy used $16$ and $100$ denoising steps.
All methods covered all tasks; the 63 sessions total 630 episodes without exclusions.

\noindent\textbf{Reproducibility and release.} We have prepared the complete real-robot deployment stack, task launchers, safety checks, and evaluation manifests for public release. These materials will be released with the training code.

\begin{table}[!htb]
\caption{Real-robot success on five tasks (three checkpoints per arm; nominally $10$ episodes per cell; three accidental repeats pooled).}
\label{tab:real_robot}
\centering
\footnotesize
\setlength{\tabcolsep}{4pt}
\begin{tabular}{@{}llcccc@{}}
\toprule
Task & Steps & DP-100 & DP-16 & MIP & \textbf{JEPA Policy} \\
\midrule
\multirow{4}{*}{Cabinet}
 & 60k  & 0/10 & 0/10 & 2/10 & \textbf{5/10} \\
 & 100k & 0/10 & 0/10 & 3/10 & \textbf{6/10} \\
 & 140k & 0/10 & 0/10 & 4/10 & \textbf{7/10} \\
 & all  & 0\% & 0\% & 30\% & \textbf{60\%} \\
\midrule
\multirow{4}{*}{Cup Stack}
 & 60k  & 1/10 & 8/20 & \textbf{8/10} & 6/10 \\
 & 100k & 1/10 & 5/20 & 2/10 & \textbf{7/10} \\
 & 140k & 1/10 & 3/10 & 2/10 & \textbf{4/10} \\
 & all  & 10\% & 32\% & 40\% & \textbf{57\%} \\
\midrule
\multirow{4}{*}{Cup Upright}
 & 60k  & 1/10 & 2/10 & 4/10 & \textbf{5/10} \\
 & 100k & 5/10 & 3/10 & \textbf{8/10} & 14/20 \\
 & 140k & 2/10 & 7/10 & 6/10 & \textbf{10/10} \\
 & all  & 27\% & 40\% & 60\% & \textbf{72\%} \\
\midrule
\multirow{4}{*}{Plate Grape}
 & 60k  & 0/10 & 0/10 & \textbf{5/10} & 2/10 \\
 & 100k & 1/10 & 0/10 & 5/10 & 5/10 \\
 & 140k & 2/10 & 3/10 & 5/10 & \textbf{7/10} \\
 & all  & 10\% & 10\% & \textbf{50\%} & 47\% \\
\midrule
\multirow{4}{*}{Pen Insert}
 & 60k  & 4/10 & 6/10 & 8/10 & \textbf{10/10} \\
 & 100k & 3/10 & 8/10 & \textbf{10/10} & \textbf{10/10} \\
 & 140k & 7/10 & 9/10 & \textbf{10/10} & 9/10 \\
 & all  & 47\% & 77\% & 93\% & \textbf{97\%} \\
\midrule
\multicolumn{2}{@{}l}{Pooled (episode-weighted)} & 18.7\% & 31.8\% & 54.7\% & \textbf{66.9\%} \\
\bottomrule
\end{tabular}

\vspace{2pt}
\parbox{\columnwidth}{\footnotesize DP-100 and DP-16 are Diffusion Policy at $100$ and $16$ denoising steps. The same two-arm platform runs every task; Cup Stack, Cup Upright and Pen Insert are performed with the right arm alone, Cabinet and Plate Grape with both. Three conditions were repeated by accident and both sessions are pooled, which is why those cells report $20$ episodes: Cup Stack DP-16 at $60$k and $100$k, and Cup Upright JEPA Policy at $100$k.}
\end{table}

\textbf{The pooled ordering matches the simulation results.} Table~\ref{tab:real_robot} reports every cell. Episode-weighted success is $66.9\%$ for JEPA Policy, $54.7\%$ for MIP, $31.8\%$ for DP-16, and $18.7\%$ for DP-100.
Across matched task--checkpoint cells, JEPA Policy leads DP-16 in $14$ of $15$ cells and DP-100 in all $15$. It is never behind either configuration. The mean margins are $+35$ and $+48$ points, with two-sided sign-test values of $p=1.22\times10^{-4}$ and $6.10\times10^{-5}$.
Because checkpoints within a task are correlated, these tests describe consistency rather than independent-task inference. Against MIP, JEPA Policy leads in $9$ of $15$ cells, trails in $4$, and ties in $2$. The mean margin is $+12.0$ points and is not significant (sign test $p=0.27$).
The cell counts should therefore not be read as 15 independent replications of each method.
They show whether the ordering persists across the three reported checkpoints, while the five tasks remain the broader units of variation.
The margin over Diffusion Policy is larger than the $+7.9$ points observed in simulation, where one task accounts for much of the difference. The margin over MIP has the same direction as the simulated $+5.6$-point gain, but its task distribution differs. It is concentrated in Cabinet ($+30.0$), while Plate Grape favors MIP ($-3.3$) and Pen Insert is nearly tied ($+3.3$, with both above $90\%$).

\textbf{Cabinet separates the arms by failure stage.} Neither Diffusion Policy configuration completes a Cabinet episode, but the failures occur at different stages.
We define the final stage as at least $194$ issued actions, the lower quartile among successful episodes. Under this definition, $7$ of $30$ DP-16 failures and $2$ of $30$ DP-100 failures placed the object but did not close the door. Video inspection confirms these outcomes (Fig.~\ref{fig:real_tasks}); the remaining failures occurred during approach or grasping.
Door closure is required for every arm. Seven of 21 MIP failures also stopped at this stage, and one successful MIP episode required three closing attempts. None of the 12 JEPA Policy failures occurred at the closing stage. Whenever JEPA Policy reached the door, it closed it.
Plate Grape shows a similar pattern. Using a task-specific threshold of $254$ actions, $3$ of $15$ MIP failures reached the final stage, compared with $0$ of $16$ for JEPA Policy and $0$ of $27$ for DP-16. DP-16 never reached the transfer stage on this task.

\section{Discussion}

The experiments indicate that future supervision within the action-generating stack improves manipulation success with little added latency. JEPA Policy adds $0.29$\,ms ($2.2\%$) to the two-pass MIP baseline. Its $13.2$\,ms decision time is $33\times$ lower than that of Diffusion Policy in the evaluated configuration (Table~\ref{tab:latency}).
The shared-versus-separate ablation provides the most direct evidence for the proposed topology. Applying the same loss through a separate branch, with or without cross-attention, provides little or no benefit. The gain therefore depends on how the future objective is integrated rather than on the mere presence of an auxiliary loss.
The Dual-independent arm reproduces the property of ACT-JEPA~\cite{act_jepa} on which this test turns: a shared conditioning path trained by both losses, feeding two mutually invisible branches. Holding target modality, anti-collapse mechanism and training recipe fixed leaves topology as the only changed variable.

The scope of these claims is limited by the observed variation across tasks.
JEPA Policy leads Diffusion Policy on all nine tasks, but Tool Hang accounts for much of the mean difference. Without Tool Hang, the mean difference is $+5.4$ points. We therefore interpret the accuracy result as comparable or better performance under the evaluated settings. The clearer advantage is computational: the measured latency differs by a factor of $33$.
The collapse audit supports a narrower statement. We observe no complete collapse or systematic contraction to the pre-specified boundary, although some tasks show low-rank exceptions. In the one-task $\lambda_a=0$ control, removing the action loss produces a boundary violation of about $36\times$. This result provides direct evidence that the action loss constrains the representation in that setting. The SIGReg~\cite{lejepa} experiment further shows that the tested explicit regularizer can severely reduce policy performance.
The rollout audit shows the future branch carries real action-outcome information, but only as a \emph{per-task-calibrated} risk signal. Raw scores do not transfer across tasks, and post-hoc thresholds remain descriptive.

Limitations.
Most experiments are conducted in simulation, and the hardware study is smaller. It covers five tasks, three checkpoints, ten episodes per cell, one robot, and one operator in one laboratory.
Individual cells are noisy at this sample size. One condition that was repeated by accident returned $5/10$ and $9/10$ for the same checkpoint and scene. The comparison with Diffusion Policy is consistent across cells ($29$ of $30$ ahead and none behind), whereas the advantage over MIP is only directional. Generalization beyond the three evaluated benchmark suites remains untested.
Table~\ref{tab:latency} isolates model time on one PPU, task, and seed with different PyTorch runtimes; RTX 3090 logs add shared-system end-to-end latency including sensing and middleware.
Like-for-like comparison against few-step action generators, distilled~\cite{consistency_policy} or single-stage~\cite{shortcut_models}, remains open.
The best-checkpoint protocol, standard in this literature, is optimistic by construction, although Table~\ref{tab:protocol} shows the reported gain survives three selection-free protocols. Single-seed side studies are directional only, and multi-modal futures remain untested.

\section{Conclusion}
JEPA Policy couples expert actions with their demonstrated future representations in a shared Transformer and predicts both in two diffusion-free passes.
Across nine simulated tasks, it improved the action-only MIP mean by $5.6$ points and exceeded both MIP and the evaluated Diffusion Policy configurations on every task. It also reduced measured decision latency from $439.5$ to $13.2$\,ms.
Ablations attributed the gain to the shared stack, and the 84-checkpoint audit found no complete collapse with action supervision.
The 5{,}200-episode rollout audit supported a per-task-calibrated rather than universal risk signal, while the five-task, 630-episode hardware study preserved the pooled method ordering.
Future work should broaden real-robot coverage, compare matched few-step generators~\cite{consistency_policy,shortcut_models}, validate diagnostic thresholds on held-out rollouts, and quantify multi-modal futures.

\begingroup
\renewcommand{\baselinestretch}{0.85}\selectfont
\bibliographystyle{IEEEtran}
\bibliography{IEEEabrv,references}
\endgroup

\end{document}